%% file: main.tex
\documentclass{article}

\usepackage{microtype}
\usepackage{graphicx}
\usepackage{subcaption}
\usepackage{booktabs} 

\usepackage[table]{xcolor}
\definecolor{bestblue}{RGB}{185, 215, 250}   
\definecolor{secondblue}{RGB}{224, 238, 255} 
\usepackage{hyperref}

\newcommand{\modelname}{\textit{WorldMirror}}
\input{math_commands.tex}

\usepackage[accepted]{icml2026}

\usepackage{amsmath}
\usepackage{amssymb}
\usepackage{mathtools}
\usepackage{amsthm}
\usepackage{multirow} 

\newcommand{\reffig}[1]{\textcolor{black}{Fig.~\ref{fig:#1}}}

\usepackage[capitalize,noabbrev]{cleveref}

\theoremstyle{plain}

\theoremstyle{definition}

\theoremstyle{remark}

\usepackage[textsize=tiny]{todonotes}

\usepackage{xcolor}
\definecolor{linkpink}{HTML}{FF007F}

\icmltitlerunning{WorldMirror: Universal 3D World Reconstruction with Any-Prior Prompting}

\makeatletter
\stepcounter{@affiliationcounter}
\newcounter{@affilzju}%
\stepcounter{@affiliationcounter}
\newcounter{@affilcuhk}%
\stepcounter{@affiliationcounter}
\newcounter{@affiltencent}%
\begin{document}

\twocolumn[
  \icmltitle{WorldMirror: Universal  3D World Reconstruction with Any-Prior Prompting}



  \icmlsetsymbol{equal}{*}

  \begin{icmlauthorlist}
    \icmlauthor{Yifan Liu}{equal,cuhk,tencent}
    \icmlauthor{Zhiyuan Min}{equal,zju,tencent}
    \icmlauthor{Zhenwei Wang}{equal,tencent}
    \icmlauthor{Junta Wu}{tencent}
    \icmlauthor{Tengfei Wang}{tencent}
    \icmlauthor{Yixuan Yuan}{cuhk}
    \icmlauthor{Yawei Luo}{zju}
    \icmlauthor{Chunchao Guo}{tencent}
  \end{icmlauthorlist}

  \icmlaffiliation{cuhk}{CUHK}
  \icmlaffiliation{tencent}{Tencent}
  \icmlaffiliation{zju}{ZJU}

  \icmlcorrespondingauthor{Tengfei Wang}{tengfeiwang12@gmail.com}
  \icmlcorrespondingauthor{Yixuan Yuan}{yxyuan@ee.cuhk.edu.hk}
  \icmlcorrespondingauthor{Yawei Luo}{yaweiluo@zju.edu.cn}

  \icmlkeywords{Machine Learning, ICML}

  \icmlteaser{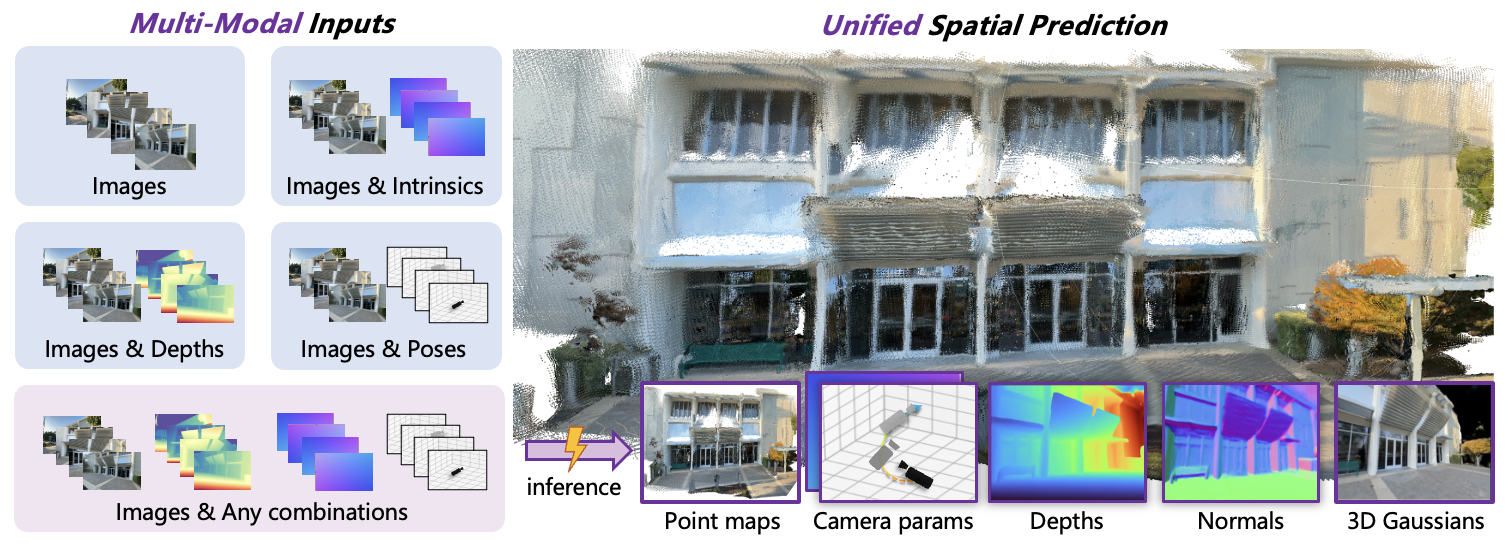}{%
    \modelname~ is a feed-forward 3D reconstruction model that takes images with optional priors (depth, intrinsics, poses) and produces point clouds, 3DGS, cameras, depth, and normals in seconds.
  }

  \vskip 0.3in

]




\printAffiliationsAndNotice{\icmlEqualContribution}



\begin{abstract}
We present \modelname, a unified feed-forward model for comprehensive 3D geometric prediction tasks. 
Unlike existing methods constrained to image-only inputs or customized for a specific task, our framework flexibly integrates diverse geometric priors, including camera poses, intrinsics, and depth maps, while simultaneously generating multiple 3D representations: dense point clouds, multi-view depth maps, camera parameters, surface normals, and 3D Gaussians. Remarkably, prior injection yields universal gains across all tasks, suggesting that input flexibility and multi-task prediction are mutually reinforcing. \modelname~achieves state-of-the-art performance across diverse benchmarks from camera, point map, depth, and surface normal estimation to novel view synthesis, while maintaining the efficiency of feed-forward inference. Code and model weights are publicly available at
\href{https://github.com/Tencent-Hunyuan/HunyuanWorld-Mirror}
  {\texttt{\textcolor{linkpink}{https://github.com/Tencent-Hunyuan\allowbreak/HunyuanWorld-Mirror}}}.
  
\end{abstract}

\section{Introduction}
Visual geometry learning is fundamental to augmented reality, robotics, and autonomous navigation. Traditional SfM~\citep{schonberger2016structure} and MVS rely on costly iterative optimization. Recently, the field has shifted toward feed-forward foundation models, with models like DUSt3R~\citep{wang2024dust3r} and VGGT~\citep{wang2025vggt} demonstrating remarkable capabilities in reconstructing geometry from image pairs, videos, and multi-view inputs. Yet, compared to foundation models in other domains, these methods remain limited in scope.

Foundation models in language and 2D vision have demonstrated the power of unified architectures that handle diverse inputs and tasks within a single framework. However, no such unified model exists for 3D geometry. Current methods remain fragmented, typically assuming RGB images as the sole input and ignoring auxiliary cues such as camera intrinsics, poses, or depth measurements that are often available in practice. They also target isolated tasks such as depth estimation~\citep{yang2024depth}, point map regression~\citep{wang2024dust3r}, or camera pose prediction~\citep{wang2023posediffusion}, requiring separate models for each application. 

We argue that a foundation model for 3D geometry should possess two key properties: (1) accepting flexible inputs that leverage auxiliary cues when available, and (2) producing comprehensive geometric outputs within a single architecture. Recent works have explored these directions separately. Pow3R~\citep{jang2025pow3r} enables prior-conditioned binocular reconstruction but outputs only point maps, while VGGT~\citep{wang2025vggt} predicts multiple geometric quantities but lacks the ability to incorporate auxiliary inputs. Unifying both properties, however, is non-trivial: flexible input conditioning must generalize across diverse auxiliary modalities without entangling with task-specific designs, while multiple outputs demand careful training strategies that become more complex with heterogeneous inputs.

To address these challenges, we introduce \textit{\modelname}, a unified end-to-end framework that performs comprehensive 3D tasks while flexibly leveraging any available geometric modalities. Unlike existing approaches that either focus on prior injection or multi-task prediction in isolation, \modelname~jointly addresses both through two core designs. The first is \textbf{Multi-modal Tokenization}, which encodes all input modalities, including images, camera intrinsics, poses, and depth maps, as tokens within a unified sequence, enabling seamless integration of any available priors without architectural modifications. All tokens are then jointly fed into the transformer backbone, allowing the model to reason over heterogeneous inputs in a unified manner. The second is \textbf{Unified Spatial Prediction}, a transformer-based architecture with unified decoder heads that handles the full spectrum of tasks from camera and depth estimation to point maps, normals, and novel view synthesis, coordinated by curriculum learning. As expected, we observe that any-modal prior injection does not merely help reconstruction, but universally boosts all output predictions. This emergent synergy highlights the advantage of unifying input flexibility and multi-task prediction within a single architecture, where shared representations enable different modalities and tasks to benefit from each other.

Extensive experiments demonstrate that \modelname~achieves state-of-the-art performance across diverse benchmarks and tasks. It surpasses recent 3D reconstruction methods, such as VGGT~\citep{wang2025vggt} and $\pi^3$~\citep{wang2025pi} in point map and camera estimation, while outperforming StableNormal~\citep{ye2024stablenormal} and GeoWizard~\citep{fu2024geowizard} in surface normal prediction and significantly exceeding recent method AnySplat~\citep{jiang2025anysplat} in novel view synthesis.

We summarize our contributions as follows: (1) We present \modelname, a unified end-to-end framework for 3D geometry that jointly addresses flexible prior conditioning and comprehensive multi-task prediction within a single model. (2) We propose Multi-modal Tokenization, which treats multiple input types including RGB images, camera intrinsics, poses, and depth as tokens, enabling seamless integration of these geometric priors without architectural modifications.  (3) We introduce a Unified Spatial Prediction architecture with a decoupled sequential training that effectively coordinates multi-task training across camera poses, depth, normals, point maps, and novel view synthesis. (4) Extensive experiments demonstrate state-of-the-art performance across diverse benchmarks, outperforming recent methods in point map estimation, camera pose prediction, surface normal estimation, and novel view synthesis.

\begin{figure*}[t]
    \centering
    \includegraphics[width=1.0\textwidth]{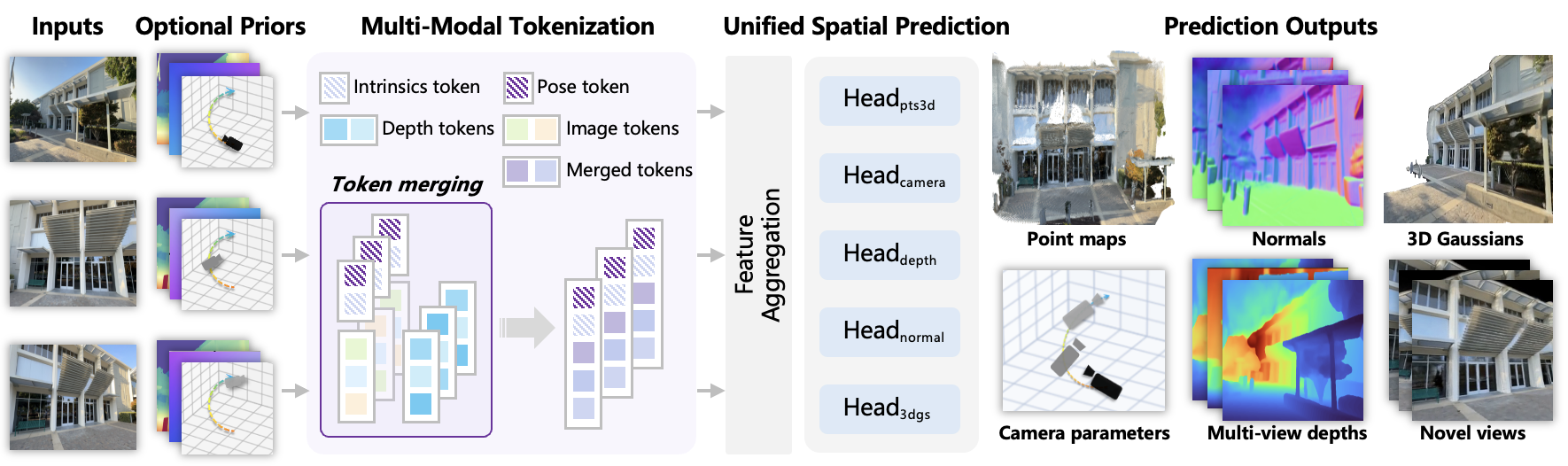}
    \caption{\textbf{Overview of \modelname.} Our framework employs \textit{Multi-modal Tokenization} to encode all inputs (images, optional priors including intrinsics, camera poses, and depth maps) into a unified token sequence. The merged tokens are processed by a transformer backbone and decoded by \textit{Unified Spatial Prediction} heads to produce comprehensive 3D outputs, including point maps, camera poses, depth maps, surface normals, and 3D Gaussians. To our knowledge, \modelname~is the first framework to unify flexible 3D inputs with comprehensive multi-task 3D prediction in a single feed-forward model.}
\end{figure*}

\section{Related Works}
\noindent\textbf{Feed-Forward 3D Reconstruction.}
Feed-forward 3D reconstruction models have emerged as powerful alternatives to traditional SfM/MVS pipelines. DUSt3R~\citep{wang2024dust3r} pioneers point map prediction, with subsequent works improving scalability~\citep{yang2025fast3r}, introducing multi-task learning~\citep{wang2025vggt}, and extending to large-scale sequences~\citep{deng2025vggt,wang2025pi}. Building on these advances, \modelname~unifies camera poses, depth, normals, point maps, and 3D Gaussians in a single feed-forward pass.


\noindent\textbf{3D Prior Guidance.}
Traditional methods like COLMAP~\citep{schonberger2016pixelwise} leverage known camera parameters for reconstruction.
Learning-based approaches explore various priors: intrinsics for monocular depth~\citep{piccinelli2024unidepth}, camera trajectories for video generation~\citep{he2024cameractrl,huang2025voyager}, and multi-modal inputs for sparse-view reconstruction~\citep{jang2025pow3r}.
Concurrently, MapAnything~\citep{keetha2025mapanything} fuses heterogeneous geometric cues with unified additive priors, while OmniVGGT~\citep{peng2025omnivggt} injects priors via per-layer zero-convolution adapters.
WorldMirror instead mixes token concatenation (global) with selective dense additions by modality, and further predicts normals and 3D Gaussians for novel view synthesis.
We study multi-modal prior injection for versatile spatial prediction under this design.

\noindent\textbf{Multi-task Learning.}
Multi-task learning (MTL) for dense prediction has been extensively studied, including uncertainty-based loss weighting~\citep{kendall2018multi}, task relationship modeling~\citep{zamir2018taskonomy}, and cross-task representation sharing~\citep{zhou2020pattern}. While joint training succeeds in 2D settings, we find that 3DGS heads (optimizing for rendering) conflict with geometry heads (optimizing for accuracy). We address this via curriculum learning: training geometry tasks first, then the GS head with frozen geometry features.

\noindent\textbf{Generalizable Novel View Synthesis.}
Novel view synthesis has advanced with NeRF~\citep{mildenhall2021nerf} and 3D Gaussian Splatting~\citep{kerbl20233d}, though these require per-scene optimization. Generalizable methods~\citep{yu2021pixelnerf, min2024entangled, charatan2024pixelsplat, min2024epipolar} enable feed-forward NVS but often rely on known camera parameters or fixed view counts. Recent pose-free approaches~\citep{ye2024no,jiang2025anysplat} eliminate calibration but with limitations in flexibility. We advance beyond these by enabling pose-free NVS with flexible view counts, optional prior injection, and superior rendering quality.

\section{Method} Given $N$ multi-view images $\{\mI_i\}_{i=1}^{N}$, our goal is to leverage any available geometric priors for unified 3D prediction. We introduce two core components: (1) \textit{Multi-modal Tokenization} (Sec.~\ref{sec:method:tokenization}), which encodes diverse input modalities, including camera intrinsics, poses, and depth maps, into a unified token sequence; and (2) \textit{Unified Spatial Prediction} (Sec.~\ref{sec:method:prediction}), a multi-task architecture with curriculum learning that produces comprehensive geometric outputs, including point maps, camera poses, depth maps, surface normals, and 3D Gaussians. We describe each component in detail below. 

\subsection{Multi-modal Tokenization}
\label{sec:method:tokenization}
A core challenge in leveraging geometric priors is flexible conditioning across diverse modalities without entangling with task-specific designs. Priors come in heterogeneous formats: intrinsics are compact matrices, poses are SE(3) transformations, and depths are dense per-pixel measurements. Rather than designing modality-specific fusion modules, we unify all inputs as tokens. Below, we describe tokenization for each modality, followed by merging and training strategies for inference with any available priors.

\noindent \textbf{Modality-Specific Tokenization.}
\textit{(i) Camera Pose:} Given poses $\{[\mR_i|\vt_i]\}_{i=1}^N$, we first normalize the scene to a unit cube via $\vt_i^{norm} = (\vt_i - \vc) / \alpha$, where $\vc$ is the scene centroid and $\alpha$ is the maximum camera-to-centroid distance, ensuring consistent numerical ranges. Each rotation $\mR_i$ is converted to a quaternion $\vq_i\in\R^{4}$, concatenated with $\vt_i^{norm}$, and projected to a pose token $\mT_i^{cam}\in \R^{1\times D}$ for each view via a two-layer MLP.

\textit{(ii) Camera Intrinsics:} Given $\mK_i\in\R^{3\times3}$, we extract the focal lengths and principal points $(f_x, f_y, c_x, c_y)$ and normalize them by the image width $W$ and height $H$. This ensures training stability across varying resolutions. The normalized 4D vector is then projected to an intrinsic token $\mT_i^{intr}\in\R^{1\times D}$ via a two-layer MLP.

\textit{(iii) Depth Map:} Unlike poses and intrinsics, depth maps are dense signals requiring a different strategy. Given $\mD_i\in\R^{H\times W}$, we normalize to $[0, 1]$ and apply a patch embedding layer to produce depth tokens $\mT_i^{depth}\in\R^{(H_p\times W_p)\times D}$ spatially aligned with visual tokens. Rather than concatenating, we add depth tokens directly to visual tokens, fusing appearance and geometry while preserving spatial structure.

\begin{figure*}[t]
    \centering
    \includegraphics[width=0.9\textwidth]{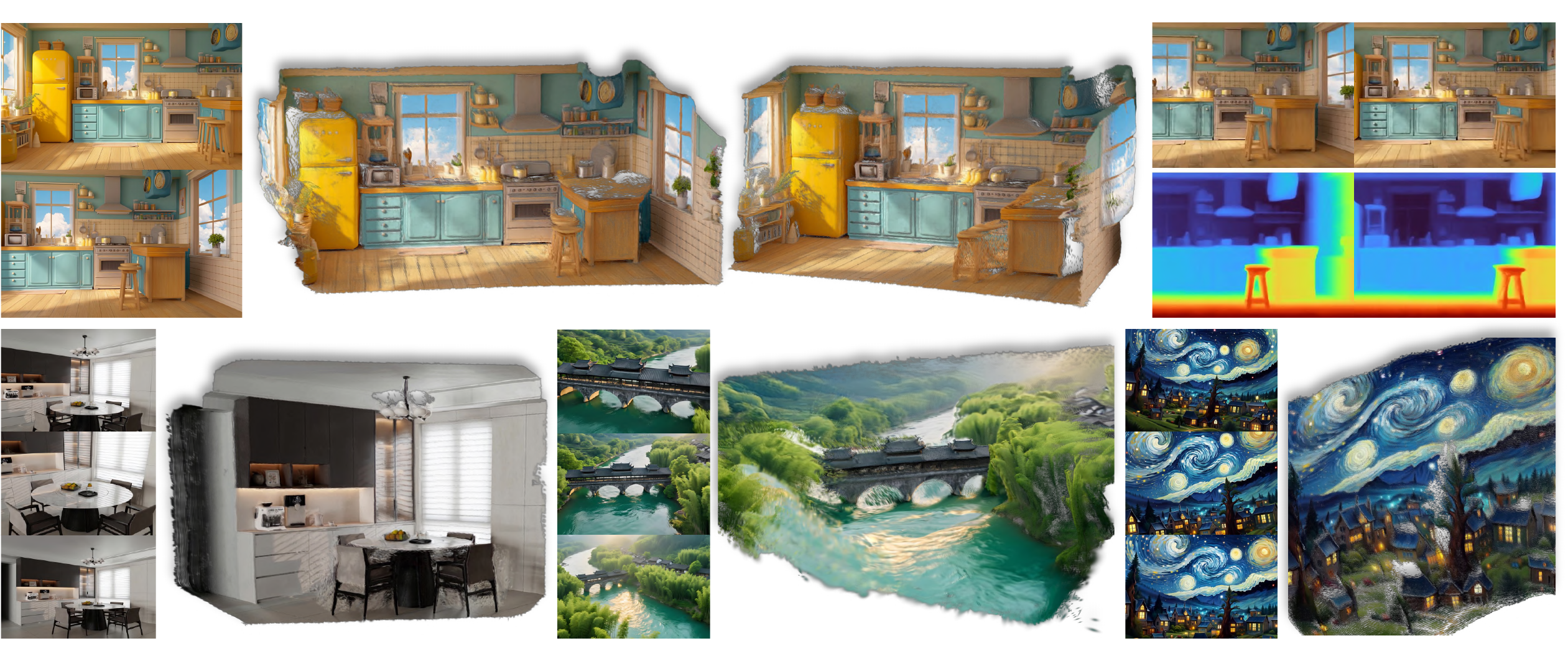}
    \caption{\textbf{Feed-Forward 3D Gaussians Predicted by~\modelname~ with In-The-Wild Inputs.} Besides real photos, our method generalizes well to AI-created videos spanning diverse styles.}
    \label{fig:ffgs_vis}
\end{figure*}

\noindent \textbf{Flexible Token Merging.} We merge all modality tokens into a unified sequence: pose and intrinsic tokens are concatenated with image tokens $\mT_i^{img}\in\R^{(H_p\times W_p)\times D}$, while depth tokens are added element-wise:
\begin{equation}
    \mT^{prompt}_i=[\mT_i^{cam},~\mT_i^{intr}, \mT_i^{img}+\mT_i^{depth}].
\end{equation}
During training, we randomly drop each of $\mT_i^{cam}$, $\mT_i^{intr}$, and $\mT_i^{depth}$ independently with probability 0.5, setting the dropped tokens to zero. This enables flexible control over input modalities at inference time.

\noindent \textbf{Remark.} We treat modalities differently based on their nature. Poses and intrinsics are global properties, making concatenation natural. Depth maps, however, are spatially dense; concatenating them would double the token count and incur quadratic attention cost. Besides, the proposed design is task-agnostic: the same merged sequence feeds all prediction heads without modality-specific branches. This agnostic property also makes future extensions straightforward, as new modalities (e.g., optical flow, semantic masks) can be incorporated by simply adding corresponding tokenizers without modifying the architecture. Finally, we observe that incorporating priors effectively improves reconstruction quality, as shown in Fig.~\ref{fig:vis-prior} and Sec. Sec.~\ref{sec:method:prediction}.

\subsection{Unified Spatial Prediction}
\label{sec:method:prediction}
To achieve a truly unified multi-task framework, we design a comprehensive architecture that jointly predicts point maps, camera parameters, depth maps, surface normals, and 3D Gaussians within a single feed-forward pass. However, jointly optimizing these tasks poses significant challenges: the entanglement of geometry and appearance learning often limits individual task performance. To address these issues, we introduce a decoupled modeling strategy that separates geometry prediction from appearance reconstruction, along with a curriculum learning scheme that progressively balances task difficulties during training.

\noindent\textbf{Geometry Modeling.}
Inspired by the architecture  used in VGGT~\citep{wang2025vggt}, we construct a Transformer backbone with a global–local attention mechanism and multi-head decoders for multi-task regression. The input images along with optional priors are tokenized as described in Sec.~\ref{sec:method:tokenization}, and the resulting tokens $\mT_i$ are fed into the backbone to extract multi-view features $\mF_i$. These features are subsequently passed to DPT decoders~\citep{ranftl2021vision} to produce dense predictions, including 3D point maps $\hat{\mP}_i$, multi-view depth maps $\hat{\mD}_i$, and surface normal maps $\hat{\mN}_i$. Additionally, $\mF_i$ is processed by an MLP decoder to estimate camera parameters $\hat{\mE}_i$.

For the surface normal estimation task, which is not included in VGGT, we apply L2 normalization to ensure unit-length vector outputs:
\begin{equation}
    \hat{\mN}_i=\mathtt{DPT}_n(\hat{\mT}^{img}_i)~/~||\mathtt{DPT}_n(\hat{\mT}^{img}_i)||_2.
\end{equation}

To address the scarcity of ground-truth normal annotations, we leverage both annotated normal labels and pseudo normals derived from ground-truth depth maps via plane fitting. This strategy enables effective utilization of diverse datasets for improved generalization.

\input{Tables/pointmap}
\input{Tables/camera}

\noindent\textbf{Appearance Modeling.}
We model appearance using 3D Gaussian Splatting to render high-quality novel view images. A special DPT decoder $\mathtt{DPT}_{g}(\cdot)$ is used to regress pixel-wise 3DGS attributes, including position $x_{g}$, color $c_{g}$, opacity $\sigma_{g}$, scale $s_{g}$, and rotation $r_{g}$. The Gaussian positions are computed from the predicted depths $\hat{\mD}_{g}$ and ground-truth camera parameters $[\mR|\vt]$. Gaussian colors are obtained by combining the original image RGB values with a predicted RGB residual. To reduce Gaussian redundancy caused by overlapping regions across multiple views, we cluster and prune per-pixel Gaussians through voxelization, similar to AnySplat~\citep{jiang2025anysplat}.

Simultaneously predicting Gaussian attributes, point cloud positions, and camera parameters is inherently challenging, as it constitutes an ill-posed problem. To mitigate the entanglement between geometric and appearance information, we carefully design $\mathtt{DPT}_{g}(\cdot)$ and the novel-view training strategy. Specifically, we apply \textbf{dual rendering supervision} on both input and novel views, compelling the model to learn a geometrically consistent 3D representation across viewpoints and effectively suppress floating artifacts. To prevent the accumulation of camera errors from corrupting appearance supervision, 3DGS rendering relies on ground-truth camera parameters rather than predicted ones. Moreover, the GS head predicts Gaussian positions independently instead of reusing outputs from the depth or point map heads, enabling the rendering task to autonomously balance geometric accuracy and appearance quality without degrading other tasks. Finally, we find that these carefully designed components stabilize training and significantly enhance rendering performance, as shown in Tab.~\ref{tab:ablation-gs}.

\noindent \textbf{Decoupled Sequential Training.}
During training, we employ a systematic decoupled sequential training designed to optimize training efficiency and enhance performance by progressing from simple to complex across task sequencing. Specifically, we jointly train the Multi-modal Tokenization module with other parameters initialized from the pretrained weights of VGGT, which establishes a foundational capability of prior-aware prediction. We then incorporate the normal prediction task into the joint training scheme. Finally, we freeze all model parameters and exclusively train the 3DGS head for 3DGS attributes prediction. This progressive task sequencing strategy ensures effective training for universal geometric prediction with any prior combination. We observe that training all tasks jointly makes it difficult for the model to disentangle geometry from appearance, resulting in suboptimal performance.

\begin{figure*}[h]
    \centering
    \includegraphics[width=0.95\textwidth]{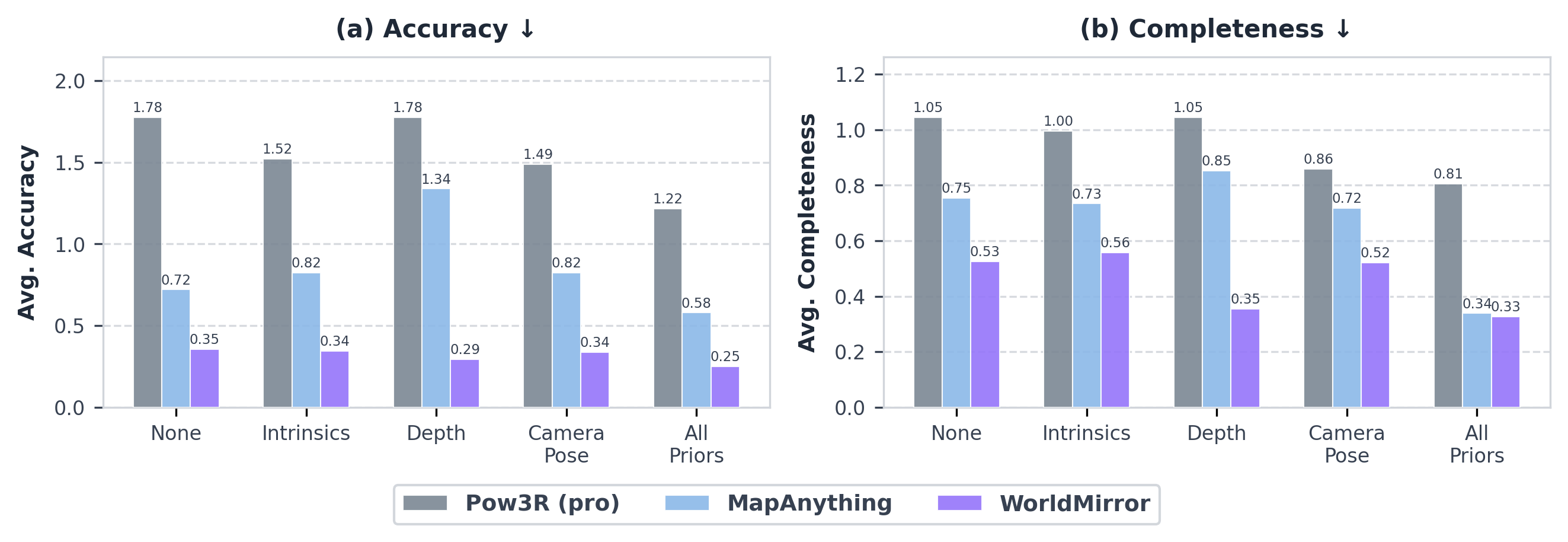}
    \caption{\textbf{Comparison with Pow3R and MapAnything under different prior conditions}. Results are averaged on 7-Scenes, NRGBD, and DTU datasets. Pow3R (pro) refers to the original Pow3R with Procrustes alignment.}
    \label{fig:comparison_pow3r_mapanything}
\end{figure*}

\begin{figure*}[h]
    \centering
    \includegraphics[width=0.9\textwidth]{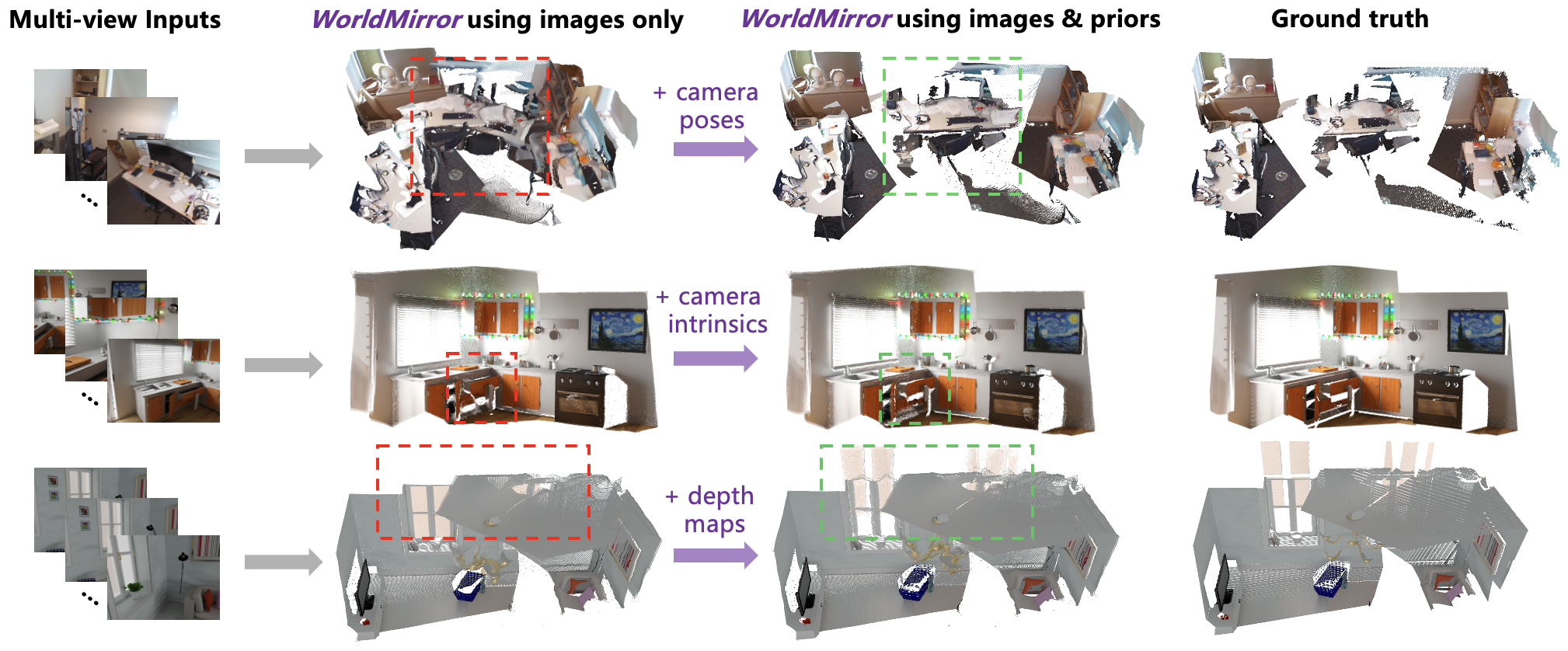}
    \caption{\textbf{Geometric Priors Unlock Enhanced Scene Reconstruction of \modelname.} (\texttt{Top}) Camera poses help the model to capture relative view positions accurately. (\texttt{Middle}) Calibrated intrinsic enhances the reconstruction by enabling precise projection modeling and geometry alignment. (\texttt{Bottom}) Depth guidance enables the network to better handle unusual geometric configurations.}
    \label{fig:vis-prior}
\end{figure*}

\noindent \textbf{Remark.}
Benefiting from our unified multi-task architecture, \modelname~achieves a true "one-for-all" paradigm where a single feed-forward pass simultaneously predicts a comprehensive spectrum of heterogeneous 3D representations, including point maps, depth, surface normals, camera parameters, and 3D Gaussians. This eliminates the need for cascaded pipelines or task-specific models, greatly simplifying deployment while maintaining competitive performance across all tasks.
Moreover, the multi-modal tokenization design enables priors from each input modality to not only enhance their respective tasks but also collaboratively improve all geometric and rendering objectives. As illustrated in Fig.~\ref{fig:vis-prior} and Tab.~\ref{fig:comparison_pow3r_mapanything}, we find that the gain arises from the 3D geometric consistency constraints imposed by the universal spatial representation.

\input{Tables/normal}
\input{Tables/nvs_multi_resolution}
\input{Figs/NVS_result1}

\section{Model Training}
Our model is trained end-to-end by minimizing a composite loss function $\Ls$ that integrates supervision for all prediction tasks:
\begin{equation}
    \Ls = \lambda_{\text{1}} \Ls_{\text{points}} + \lambda_{\text{2}} \Ls_{\text{depth}} + \lambda_{\text{3}} \Ls_{\text{cam}} + \lambda_{\text{4}} \Ls_{\text{normal}} + \lambda_{\text{5}} \Ls_{\text{3dgs}}
\end{equation}
where $\Ls_{\text{points}}$ supervises point map regression, $\Ls_{\text{depth}}$ supervises multi-view depth estimation, $\Ls_{\text{cam}}$ supervises camera pose prediction, $\Ls_{\text{normal}}$ supervises surface normal estimation, and $\Ls_{\text{3dgs}}$ supervises 3D Gaussian prediction through rendering losses. Please refer to Sec. \ref{sec:training_losses} for the details of training losses and the specific values of these weights.

\section{Experiments}
We evaluate \modelname~on comprehensive tasks: point map reconstruction, camera pose estimation, surface normal estimation, and novel view synthesis (\S\ref{sec:exp:benchmark}). We further assess prior-guidance effectiveness (Sec. \ref{sec:exp:prior}) and conduct ablations on prior embedding strategies and novel view synthesis design choices (Sec. \ref{sec:exp:ablation}). Qualitative results on AI-generated videos demonstrate in-the-wild generalization (\reffig{ffgs_vis_add}, \reffig{pointmap_wogt}). Training details are in Sec. \ref{sec:training_settings}.

\subsection{Evaluation on Different Tasks}
\label{sec:exp:benchmark}

\noindent\textbf{Point Map Reconstruction.}
We evaluate point map reconstruction on scene-level datasets, including 7-Scenes~\citep{shotton2013scene}, NRGBD~\citep{azinovic2022neural} and object-level  dataset DTU~\citep{jensen2014large}, using the same sequence mappings as~\citep{wang2025pi}. As shown in Tab.~\ref{tab:pointmap}, our method without priors already surpasses VGGT and $\pi^3$, with 10.4\% and 17.8\% accuracy gains on 7-Scenes and DTU. Incorporating priors further improves results; using all priors yields 58.1\% and 53.1\% accuracy gains on 7-Scenes and NRGBD over our no-prior baseline, demonstrating effective prior utilization.

\noindent\textbf{Camera Pose Estimation.}
Following the protocol of~\citep{wang2025pi}, we evaluate camera pose estimation on RealEstate10K~\citep{zhou2018stereo}, Sintel~\citep{bozic2021transformerfusion}, and TUM-dynamics~\citep{sturm2012benchmark}. We report RRA, RTA, and AUC@30° for RealEstate10K, and ATE, RPE for Sintel and TUM-dynamics. As shown in Tab.~\ref{tab:camera_pose}, our method achieves superior zero-shot performance on RealEstate10K and TUM-dynamics, while remaining competitive on Sintel despite limited outdoor dynamic scenes data involved in the training datasets.

\input{Tables/ablation_prior}
\input{Tables/abaltion_gs}

\noindent\textbf{Surface Normal Estimation.}
Following~\citep{bae2024rethinking}, we evaluate on iBims-1~\citep{koch2018evaluation}, NYUv2~\citep{silberman2012indoor}, and ScanNet~\citep{dai2017scannet}, reporting mean/median angular error and percentage below 22.5°/30.0° thresholds. Tab.~\ref{tab:normal_estimation} shows substantial improvements over existing methods, demonstrating that multi-task learning with shared representations can outperform specialized single-task approaches.

\noindent\textbf{Novel View Synthesis.}
We introduce a multi-resolution benchmark on DL3DV~\citep{ling2024dl3dv} following FLARE~\citep{zhang2025flare} to fairly compare methods with different input resolutions (details in Sec.~\ref{sec:details_of_nvs}). As shown in Tab.~\ref{tab:dl3dv_multiresolution_depthsplat}, our method consistently outperforms DepthSplat~\citep{xu2025depthsplat} at 8, 24, and 64 views, with gains increasing as view count grows. This improvement likely stems from a combination of our unified architecture and the broader training data scale, where the expanded data coverage provides richer scene diversity and contributes to the observed gains. While acknowledging this data advantage, our unified design still better exploits multi-view correlations as more observations become available. Trained with dynamic resolutions, our model generalizes robustly across varying resolutions and consistently surpasses baselines. Additional results on post-optimization, more input views, and two-view settings are provided in Sec.~\ref{sec:nvs_optimization}, Sec.~\ref{sec:nvs_matrixcity}, and Sec.~\ref{sec:nvs_two-view}.

\subsection{Evaluation on Different Input Configurations}
\label{sec:exp:prior}
Fig.~\ref{fig:vis-prior} illustrates how different priors contribute to reconstruction quality, with quantitative results in Sec.~\ref{appendix:Different Input Configurations}. Camera poses capture global geometry, calibrated intrinsics resolve scale ambiguity, and depth priors provide pixel-level constraints for complex regions. Combining all priors yields further gains, confirming that multi-modal priors work synergistically with complementary geometric cues.

\subsection{{Comparison with Prior-guided Methods}}
We also compare with recent prior-guided methods Pow3R~\citep{jang2025pow3r} and MapAnything~\citep{keetha2025mapanything} under different prior conditions. Note that Pow3R was designed for 2-view reconstruction; we extend it to multi-view scenarios using Procrustes alignment for camera pose estimation. Results are shown in Fig.~\ref{fig:comparison_pow3r_mapanything}. As demonstrated, \modelname~consistently outperforms both methods across most conditions. Compared to Pow3R, our method employs more multi-view-friendly embedding strategies that better preserve geometric consistency. Compared to MapAnything, our model benefits from fine-tuning on VGGT, leading to improved generalization on out-of-domain data.

\subsection{Ablation Study}
\label{sec:exp:ablation}
\noindent\textbf{Prior Embedding Ablation.} 
We explore different ways of embedding priors in Tab.~\ref{tab:ablation-prior}.
For camera poses, we experiment with (1) dense Plücker ray embeddings that are added element-wise to the image tokens, and (2) a single token concatenation approach where the pose is compressed into a single token and concatenated to the sequence. For camera intrinsics, we similarly compare dense raymap embeddings that are added to the image tokens versus a single token. Our experiments reveal that the single token approach achieves better performance for embedding both camera poses and intrinsics, suggesting that a compact global representation is more effective than dense per-pixel conditioning while being more efficient.

\noindent\textbf{Novel View Synthesis Ablation.}  Tab.~\ref{tab:ablation-gs} reports ablation analysis on novel view synthesis: (1) We replace ground-truth camera parameters with predicted ones for 3DGS rendering to examine their importance. (2) We remove novel-view rendering supervision similar to~\citep{jiang2025anysplat} to assess the necessity of supervising both input and novel views. (3) The GS head predicts all Gaussian attributes except positions, which are derived from depth maps estimated by the Depth head. These studies confirm that both our 3DGS prediction framework and training strategy are essential for optimal performance.

\section{Conclusion}
We presented \modelname, a unified end-to-end framework addressing flexible prior conditioning and comprehensive multi-task prediction for 3D geometry. This design fills a gap where prior methods address these capabilities in isolation. Through Multi-modal Tokenization, our model integrates images, camera intrinsics, poses, and depth maps as tokens without architectural modifications. Coordinated by curriculum learning, Unified Spatial Prediction handles tasks from camera estimation to novel view synthesis, where prior injection universally boosts all predictions. Experiments show \modelname~achieves state-of-the-art across point maps, camera poses, surface normals, and novel view synthesis. This establishes unified, prior-aware architectures as a promising direction for versatile 3D understanding.

\section*{Acknowledgment}
This work was support by CUHK Start-up Grant, CUHK SSFCRS 23/24,  National Natural Science Foundation of China (62293554, U2336212), ``Pioneer'' and ``Leading Goose'' R\&D Program of Zhejiang (2024C01073), Ningbo Innovation ``Yongjiang 2035'' Key Research and Development Programme (2024Z292), and Natural Science Foundation of Zhejiang Province, China (LZ24F020002).

\section*{Impact Statement}
\modelname~enables efficient 3D scene reconstruction across multiple applications, making advanced geometric prediction more accessible. While benefiting researchers and small teams, we recognize concerns about privacy implications, potential misrepresentation of environments, and algorithmic biases. We encourage ongoing research into verification methods and ethical guidelines for responsible implementation of 3D reconstruction technologies.

\nocite{langley00}
\bibliography{icml2026_conference}
\bibliographystyle{icml2026}

\newpage
\appendix
\onecolumn


\section{Model Training}
\label{sec:model_training}
\subsection{Training Losses} 
\label{sec:training_losses}
Our model is trained end-to-end by minimizing a composite loss function, $\Ls$, which integrates supervision for all prediction tasks:
\begin{equation}
    \Ls = \lambda_{\text{points}}\Ls_{\text{points}} + \lambda_{\text{depth}}\Ls_{\text{{depth}}} + \lambda_{\text{cam}}\Ls_{\text{{cam}}} + \lambda_{\text{normal}}\Ls_{\text{{normal}}} + \lambda_{\text{3dgs}}\Ls_{\text{{3dgs}}}.
\end{equation}
We follow VGGT to implement $\Ls_{\text{{cam}}}$, $\Ls_{\text{points}}$, and $\Ls_{\text{{depth}}}$. Specifically, we use  a gradient-based term to supervise the predicted point $\hat{P}_i$: 
\begin{equation}
    \Ls_{\text{point}} = \sum_{i=1}^{N} \|\Sigma_i^P \odot (\hat{\mP}_i - \mP_i)\| + \|\Sigma_i^P \odot (\nabla\hat{\mP}_i - \nabla \mP_i)\| - \alpha \log \Sigma_i^P\label{eq:7},
\end{equation}
where $\odot$ is the channel-broadcast element-wise product and $\Sigma_i^P$ refers to the point uncertainty. The depth loss $\Ls_{\text{depth}}$ is analogous to $\Ls_{\text{point}}$ but replaces the point with depth. For camera loss $\Ls_{\text{cam}}$, we implement a Huber loss $\|\cdot\|_{\epsilon}$ to supervise the predicted camera $\mE_i$: 
\begin{equation}
    \Ls_{\text{cam}}=\Sigma_{i=1}^N\|\mE_i-\hat{\mE}_i\|_{\epsilon}.
\end{equation}
To supervise the predicted surface normals $\hat{\mE_i}$, we use Angle Loss (AL), which effectively measures the directional deviation between predicted and ground truth normal vectors. The normal loss function is specifically defined as:
\begin{equation}
    \mathcal{L}_{\text{normal}} = \Sigma_{i=1}^{N} \alpha_l \cdot (1 - |\hat{\mN}_i \cdot \mN_i|).
\end{equation}


To enhance robustness in novel views, at each training iteration, we partition the input views $ I $ into \( K \) candidate context and novel view splits. The pixel overlap rate between the ground truth depth map and camera parameters is computed for each novel view in the context of the candidate context views. The split with the highest pixel overlap rate is selected, with the corresponding context views and novel views being used for further training. Next, based on the selected context images, we regress the 3DGS positions and properties, and render both context view images and novel view images $ \hat{I}$. Then, the RGB rendering loss across all views is defined as follows:
\begin{equation}
\Ls_{rgb} = \Sigma_{i=1}^N\|I_i[M_i] - \hat{I}_i[M_i]\| + \lambda_{\text{lpips}} \mathtt{LPIPS}(I_i[M_i], \hat{I_i}[M_i]),
\end{equation}
where 
$M$ denotes the mask indicating whether the pixels in the current view are visible from the context views, analogous to the novel view mask introduced in~\cite{smart2024splatt3r}. 

To explicitly supervise the locations of the 3D Gaussian splats, we introduce the depth supervision loss $\Ls_{\text{gsdepth}}$, which enforces consistency between the ground truth depth map and the depth map predicted by the GS head. The formulation of $\Ls_{gsdepth}$ follows the same definition as Eq.~\ref{eq:7}. It is worth noting that, instead of using the depth estimated by the depth head to compute the Gaussian positions, we rely on the GS head to directly predict both the positions and other attributes of the splats. This design choice is further validated in our ablation studies (see Tab.~\ref{tab:ablation-gs}). However, due to inherent ambiguities in multi-view rendering and potential noise in the ground truth depth, relying solely on $\Ls_{\text{rgb}}$ and $\Ls_{gsdepth}$ often leads to the presence of floating points in the predicted 3DGS. To mitigate this issue, we introduce a gradient consistency loss $\Ls_{\text{consis}}$, which regularizes the gradients of the GS-rendered depth map $\tilde{D}$ to be consistent with the pseudo depth $\hat{D}$ predicted by the depth head:
\begin{equation}
\Ls_{\text{consis}} = \Sigma_{i=1}^N\|{\nabla\hat{D}_i}[\hat{M}_i] - \nabla\tilde{D}_i[\hat{M}_i]\|,
\end{equation}
where $\hat{M}$ is the depth confidence mask corresponding to the top $30\%$-quantile of the confidence map. Finally, the 3DGS loss is defined as $\Ls_{\text{3dgs}} = \Ls_{\text{rgb}} + \lambda_{\text{gsdepth}}\Ls_{\text{gsdepth}}+\lambda_{\text{consis}}\Ls_{\text{consis}}$.

\input{Tables/depth}

\input{Tables/nvs2}

\subsection{Training Settings}
\label{sec:training_settings}
\noindent \textbf{Implementation Details.} 
Our model undergoes a two-phase training process. Initially, we train for 100 epochs using multi-modal prior prompting with a normal head, followed by 50 epochs of fine-tuning with a Gaussian head. Throughout both phases, we implement dynamic image resolutions, maintaining total pixel counts between 100,000 and 250,000, while sampling aspect ratios from 0.5 to 2.0. We employ a dynamic batch sizing approach similar to VGGT, processing 24 images per GPU across a cluster of 32 H20 GPUs. Our optimization strategy features parameter-specific learning rates: 2e-5 for patch embedding layers, 1e-4 for alternated attention modules and pre-trained pointmap, depth, and camera head, and 2e-4 for newly introduced parameters. We use a CosineAnnealing scheduler that gradually decreases from maximum to minimum values following a cosine curve. For our composite loss function, we carefully balance component weights as follows: $\lambda_{\text{points}}=1.0$, $\lambda_{\text{depth}}=1.0$, $\lambda_{\text{cam}}=5.0$, $\lambda_{\text{normal}}=1.0$, $\lambda_{\text{3dgs}}=1.0$, $\lambda_{\text{lpips}}=0.05$, $\lambda_{\text{gsdepth}}=0.1$, $\lambda_{\text{consis}}=0.1$.

\noindent \textbf{Training Data.} 
The training data comprises a diverse collection of 15 datasets spanning various scene types and capture conditions. This heterogeneous mix includes both established benchmarks and recent collections: DL3DV~\citep{ling2024dl3dv}, BlenderMVS~\citep{yao2020blendedmvs}, TartanAir~\citep{wang2020tartanair}, ASE~\citep{pan2023aria}, Unreal4K~\citep{tosi2021smd}, Habitat~\citep{savva2019habitat}, MapFree~\citep{arnold2022map}, MVS-Synth~\citep{DeepMVS}, ArkitScenes~\citep{dehghan2021arkitscenes}, ScanNet++~\citep{yeshwanth2023scannet++}, MegaDepth~\citep{MegaDepthLi18}, Hypersim~\citep{roberts2021hypersim}, Matterport3D~\citep{chang2017matterport3d}, Co3dv2~\citep{reizenstein2021common}, and WildRGBD~\citep{xia2024rgbdobjectswildscaling}  datasets. This extensive dataset aggregation provides rich supervision across indoor/outdoor environments, real/synthetic scenes, and static/dynamic objects, enabling our model to learn generalizable geometric representations.

\section{Additional Comparisons}
\label{sec: Additional Comparisons}
\subsection{Monocular and video depth benchmark}
In Tab.~\ref{tab:depth_estimation}, we evaluate \modelname~in comparison with contemporary approaches for both single-view and sequential depth estimation across diverse input scenarios. Despite \modelname~not being explicitly optimized for monocular metric depth inference, it delivers performance that matches or exceeds current leading methods. When processing video sequences, \modelname~produces results that rival specialized feed-forward reconstruction frameworks. We note a modest performance gap on the KITTI benchmark relative to $\pi^3$, which we attribute to the under-representation of urban driving environments in our training distribution. Future iterations of our work will incorporate a more comprehensive collection of street-level imagery to enhance generalization to such scenarios.

\subsection{Implementation details for the evaluation of novel view synthesis}
\label{sec:details_of_nvs}
We evaluate zero-shot novel view synthesis on three datasets: RealEstate10K~\citep{zhou2018stereo}, DL3DV~\citep{ling2024dl3dv}, and VR-NeRF~\citep{xu2023vr} under both sparse-view and dense-view settings. For RealEstate10K, we randomly sample 200 scenes from the NopoSplat~\citep{ye2024no} test split, using 3 novel views per scene in the sparse-view setting and 4 novel views per scene in the dense-view setting. For DL3DV, we follow the FLARE test split and evaluate in 112 unseen scenes, each containing 9 novel views. For VR-NeRF, consistent with AnySplat, we select 5 scenes, each with 64 input views and 6 novel views. For calculating the rendering metrics, we follow the  \textit{test-time camera pose alignment} introduced by AnySplat to ensure fair evaluation. 

Tab.~\ref{tab:novel_view_synthesis1} reports the quantitative evaluation results at a unified resolution of $518\times378$ for novel view synthesis under the feed-forward setting. Our method achieves substantial improvements over the previous state-of-the-art AnySplat, with consistent gains across all metrics on both datasets, demonstrating the effectiveness of our unified geometric representation for high-quality view synthesis.

\input{Tables/nvs1}

\begin{table*}[t]
\centering
\caption{\textbf{Novel view synthesis results on MatrixCity~\citep{li2023matrixcity} using 100, 150, and 200 input views. }
\modelname~consistently outperforms prior feed-forward and optimization-based methods across most metrics.
}
\label{tab:matrixcity_highview_nvs}
\resizebox{0.95\textwidth}{!}{
\begin{tabular}{lccccccccc}
\toprule
\multicolumn{1}{l}{\multirow{3}{*}{{\textbf{Method}}}}
& \multicolumn{3}{c}{{\textbf{MatrixCity} (100 views)}} 
& \multicolumn{3}{c}{{\textbf{MatrixCity} (150 views)}} 
& \multicolumn{3}{c}{{\textbf{MatrixCity} (200 views)}} \\
\cmidrule(r){2-4} \cmidrule(r){5-7} \cmidrule(r){8-10}
& {PSNR $\uparrow$} 
& {SSIM $\uparrow$} 
& {LPIPS $\downarrow$}
& {PSNR $\uparrow$} 
& {SSIM $\uparrow$} 
& {LPIPS $\downarrow$}
& {PSNR $\uparrow$} 
& {SSIM $\uparrow$} 
& {LPIPS $\downarrow$} \\
\midrule

{3D-GS~\citep{kerbl20233d}}
& {18.21} & {0.568} & {0.445}
& {18.86} & {0.593} & {0.412}
& {19.54} & {0.612} & {0.388} \\

{Mip-Splatting~\citep{yu2024mip}}
& {17.97} & {0.536} & {0.450}
& {18.24} & {0.579} & {0.438}
& {18.63} & {0.588} & {0.414} \\

{AnySplat~\citep{jiang2025anysplat}}
& {20.51} & {0.620} & {\textbf{0.347}}
& {19.24} & {0.601} & {0.399}
& {19.18} & {0.605} & {0.397} \\

{\textbf{WorldMirror}}
& {\textbf{20.88}} & {\textbf{0.640}} & {0.360}
& {\textbf{20.62}} & {\textbf{0.626}} & {\textbf{0.370}}
& {\textbf{20.36}} & {\textbf{0.630}} & {\textbf{0.375}} \\

\bottomrule
\end{tabular}
}
\end{table*}

\begin{table*}[t]
\centering
\caption{\textbf{Two-view NVS comparison on RealEstate10K and DL3DV.} WorldMirror demonstrates strong generalization ability, even without being trained specifically for the two-view NVS setting.}
\label{tab:twoview_nvs_noposplat}
\resizebox{0.95\textwidth}{!}{
\begin{tabular}{lccccccc}
\toprule
\multicolumn{1}{l}{\multirow{3}{*}{{\textbf{Method}}}}
& \multicolumn{1}{l}{\multirow{3}{*}{{\textbf{Prior-Type}}}}
& \multicolumn{3}{c}{{\textbf{RealEstate10K} (2 views)}}
& \multicolumn{3}{c}{{\textbf{DL3DV} (2 views)}} \\
\cmidrule(r){3-5} \cmidrule(r){6-8}
& 
& {PSNR $\uparrow$}
& {SSIM $\uparrow$}
& {LPIPS $\downarrow$}
& {PSNR $\uparrow$}
& {SSIM $\uparrow$}
& {LPIPS $\downarrow$} \\
\midrule

{NoPoSplat~\citep{ye2024no}} 
& {Intrinsics}
& {\textbf{25.06}}
& {\textbf{0.836}}
& {0.164}
& {19.00}
& {0.575}
& {0.350} \\

{AnySplat~\citep{jiang2025anysplat}} 
& {None} 
& {18.01}
& {0.602}
& {0.207}
& {13.56}
& {0.368}
& {0.338} \\

{WorldMirror} 
& {None} 
& {23.48}
& {0.805}
& {0.124}
& {18.41}
& {0.582}
& {0.270} \\

{WorldMirror} 
& {Intrinsics}
& {23.89}
& {0.826}
& {\textbf{0.113}}
& {\textbf{19.08}}
& {\textbf{0.636}}
& {\textbf{0.250}} \\

\bottomrule
\end{tabular}
}
\end{table*}

\subsection{Novel view synthesis with optimization}
\label{sec:nvs_optimization}
Although recent feed-forward pipelines are capable of synthesizing competitive 3D Gaussian splats (3DGS) within seconds, they inevitably suffer from errors introduced by single-pass predictions, such as suboptimal Gaussian placement and appearance. We hypothesize that incorporating a brief post-optimization stage—initialized with either our predicted point cloud or 3DGS primitives—can significantly refine both geometry and appearance at only modest additional cost, thereby accelerating the convergence of 3DGS training and enhancing rendering quality. 

As shown in Tab.~\ref{tab:novel_view_synthesis2}, we compare (i) feed-forward baselines and (ii) post-optimization with 3,000 or 1,000 iterations, initialized either from a random point cloud or from feed-forward 3DGS primitives. The camera parameters for optimizing 3DGS are obtained from the feed-forward outputs of the chosen method. Our predicted point cloud, camera, and 3DGS primitives provide a robust and high-quality initialization for 3DGS optimization, significantly accelerating the training process and consistently surpassing baseline methods across all metrics.

\subsection{Novel view synthesis with large-scale multi-view inputs}
\label{sec:nvs_matrixcity}
{We evaluate novel view synthesis with 100, 150, and 200 input views on the MatrixCity dataset~\citep{li2023matrixcity}. Following AnySplat~\citep{jiang2025anysplat}, we use a resolution of $448\times448$ for all methods and compare \modelname~ with AnySplat~\citep{jiang2025anysplat}, 3D-GS~\citep{kerbl20233d}, and Mip-Splatting~\citep{yu2024mip}. As shown in the Tab.~\ref{tab:matrixcity_highview_nvs}, the experiment demonstrates that our method generalizes far beyond the maximum of 24 input views used during training, and achieves superior performance to both feed-forward and optimization-based approaches, without any post-processing or additional refinement.}

\subsection{Novel view synthesis in the two-view setting}
\label{sec:nvs_two-view}
{We conduct two-view NVS experiments on both RealEstate10K~\citep{zhou2018stereo} and DL3DV~\citep{ling2024dl3dv} compared with NoPoSplat~\citep{ye2024no}. NoPoSplat takes inputs at a resolution of $256\times256$, whereas AnySplat~\citep{jiang2025anysplat} and \modelname~use input images at $406\times406$ (center-cropping and resizing to ensure an equivalent receptive field to $256\times256$). All methods are evaluated at $256\times256$ by downsampling the rendered results. As shown in the Tab.~\ref{tab:twoview_nvs_noposplat}, our model is not trained specifically on RealEstate10K for the two-view setting, yet it achieves performance comparable to NoPoSplat across most metrics. Moreover, when compared to AnySplat, which more closely matches our training configuration, our method substantially outperforms it.}

\begin{table*}[h]
\centering
\caption{\textbf{Robustness evaluation of WorldMirror with noisy priors on 7-Scenes and DTU datasets.} The model exhibits graceful degradation under various noise conditions.}
\label{tab:noise_robustness}
\resizebox{0.8\textwidth}{!}{
\begin{tabular}{l|l|cccc}
\toprule
{\textbf{Prior Type}} & {\textbf{Noise Level}} & {\textbf{7S-Acc.} $\downarrow$} & {\textbf{7S-Comp.} $\downarrow$} & {\textbf{DTU-Acc.} $\downarrow$} & {\textbf{DTU-Comp.} $\downarrow$} \\
\midrule
{None (baseline)} & {-} & {0.044} & {0.050} & {0.982} & {1.486} \\
\midrule
\multirow{3}{*}{{Pose (Rotation)}} & {0° (clean)} & {\textbf{0.022}} & {\textbf{0.035}} & {\textbf{0.966}} & {1.502} \\
& {20°} & {0.023} & {\textbf{0.035}} & {0.976} & {1.483} \\
& {40°} & {0.025} & {0.036} & {1.017} & {\textbf{1.466}} \\
\midrule
\multirow{3}{*}{{Pose (Translation)}} & {$\sigma=0.0$ (clean)} & {\textbf{0.022}} & {\textbf{0.035}} & {\textbf{0.966}} & {1.502} \\
& {$\sigma=0.05$} & {\textbf{0.022}} & {\textbf{0.035}} & {0.967} & {1.503} \\
& {$\sigma=0.1$} & {0.024} & {0.036} & {0.967} & {1.503} \\
\midrule
\multirow{3}{*}{{Intrinsics}} & {$1.0\times$ (clean)} & {0.047} & {0.051} & {\textbf{0.948}} & {\textbf{1.579}} \\
& {$0.8\times$} & {\textbf{0.044}} & {\textbf{0.049}} & {1.047} & {1.740} \\
& {$0.6\times$} & {0.051} & {0.054} & {2.691} & {1.830} \\
\midrule
\multirow{3}{*}{{Depth}} & {$\sigma=0.0$ (clean)} & {\textbf{0.058}} & {\textbf{0.060}} & {\textbf{0.790}} & {\textbf{0.977}} \\
& {$\sigma=0.05$} & {\textbf{0.058}} & {0.061} & {0.795} & {0.980} \\
& {$\sigma=0.1$} & {0.064} & {0.071} & {1.387} & {1.820} \\
\bottomrule
\end{tabular}
}
\end{table*}

\begin{figure*}[t]
    \centering
    \includegraphics[width=1.0\textwidth]{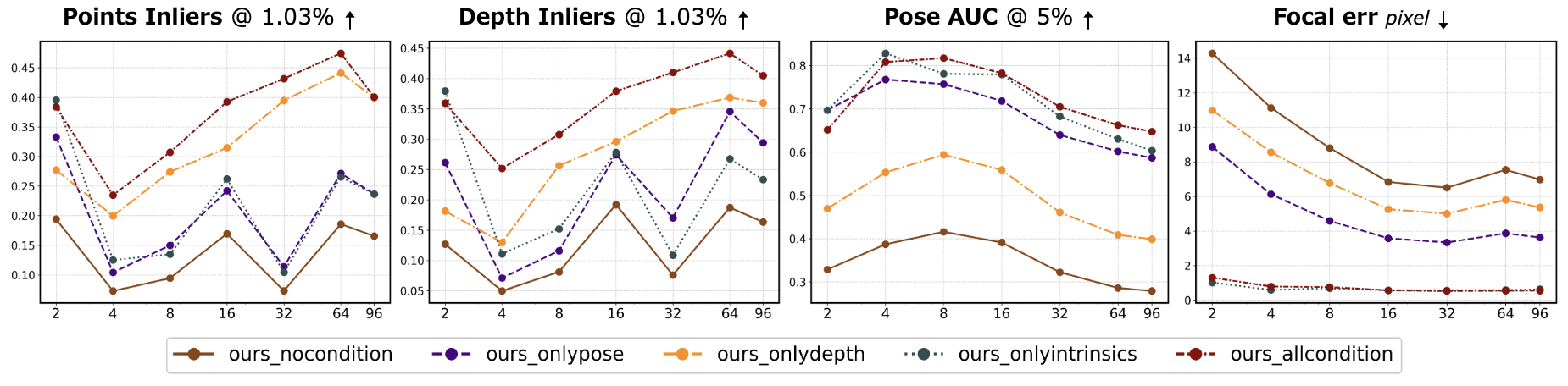}
    \caption{\textbf{Geometric Priors Boosts Model's Feed-Forward Performance across All Tasks.} Incorporating a single modality not only enhances predictions for its corresponding task but also improves performance across other tasks. This suggests that modal information enables the model to develop a more comprehensive understanding of the overall geometry.}
    \label{fig:num-prior}
\end{figure*}

\subsection{Additional experiments under different input configurations}
\label{appendix:Different Input Configurations}
To demonstrate the benefits of incorporating priors into model predictions, we evaluate model performance across various input configurations. We present four key metrics: the inlier ratio at a relative threshold of 1.03\% of points and depths, the area under the curve at a 5° error threshold (AUC@5), and the average focal error in pixels, measured across the ETH3D~\citep{schops2017multi} and DTU~\citep{jensen2014large} datasets. As shown in Fig.\ref{fig:num-prior}, incorporating even a single modality prior yields dual benefits: it enhances both the corresponding task prediction and the model's capacity to infer other geometric attributes. 

\section{Additional Ablation Studies}

\subsection{Curriculum Learning vs. Joint Training}

To validate our decoupled training strategy, we compare it against a joint training baseline in which the geometry backbone and all heads are optimized simultaneously from the outset. For a fair comparison under time constraints, both variants are trained for 32{,}500 steps on 32 GPUs. As reported in Tab.~\ref{tab:ablation_curriculum}, our decoupled strategy consistently outperforms joint training across nearly all benchmarks.

\begin{table}[h]
\centering
\caption{\textbf{Ablation study comparing our decoupled training strategy against joint training.} Lower is better for all error metrics ($\downarrow$); higher is better for PSNR ($\uparrow$). Best results are in \textbf{bold}.}
\small
\resizebox{0.99\textwidth}{!}{
\begin{tabular}{lcccccccc}
\toprule
\textbf{Strategy} & \textbf{7S-Acc $\downarrow$} & \textbf{7S-Comp $\downarrow$} & \textbf{NRGBD-Acc $\downarrow$} & \textbf{NRGBD-Comp $\downarrow$} & \textbf{DTU-Acc $\downarrow$} & \textbf{DTU-Comp $\downarrow$} & \textbf{RE10K PSNR $\uparrow$} & \textbf{DL3DV PSNR $\uparrow$} \\
\midrule
Joint Training & 0.048 & 0.077 & 0.084 & 0.085 & 1.264 & \textbf{1.792} & 20.12 & 20.31 \\
Decoupled (Ours) & \textbf{0.043} & \textbf{0.049} & \textbf{0.041} & \textbf{0.045} & \textbf{1.017} & 1.780 & \textbf{20.62} & \textbf{20.92} \\
\bottomrule
\end{tabular}
}
\label{tab:ablation_curriculum}
\end{table}

\subsection{Sensitivity Analysis on Prior Dropout Probability}

The prior dropout probability $p$ governs the proportion of training iterations in which depth and normal priors are randomly withheld from the model, thereby encouraging robustness to missing or unavailable priors at inference time. To characterize the sensitivity of our approach to this hyperparameter, we perform a systematic sweep over $p \in \{0.0, 0.3, 0.5, 0.7, 1.0\}$ and evaluate each configuration under two inference regimes: \emph{no-prior} (all priors withheld) and \emph{all-prior} (all priors provided). Results are summarized in Tab.~\ref{tab:ablation_dropout}.

\begin{table}[h]
\centering
\caption{\textbf{Sensitivity analysis on the prior dropout probability $p$.} Performance is reported under both no-prior and all-prior inference conditions. Lower is better for all accuracy metrics ($\downarrow$). Best results are in \textbf{bold}.}
\small
\begin{tabular}{lcccc}
\toprule
\textbf{Dropout $p$} & \textbf{No Prior: 7S-Acc $\downarrow$} & \textbf{No Prior: DTU-Acc $\downarrow$} & \textbf{All Priors: 7S-Acc $\downarrow$} & \textbf{All Priors: DTU-Acc $\downarrow$} \\
\midrule
0.0 (always on) & 0.068 & 1.356 & \textbf{0.015} & \textbf{0.698} \\
0.3 & 0.049 & 1.042 & 0.019 & 0.728 \\
0.5 (Ours) & \textbf{0.044} & \textbf{0.982} & 0.018 & 0.735 \\
0.7 & 0.044 & 0.989 & 0.021 & 0.753 \\
1.0 (always off) & 0.047 & 0.985 & 0.049 & 0.988 \\
\bottomrule
\end{tabular}
\label{tab:ablation_dropout}
\end{table}

\section{{Robustness to Noisy or Low-Quality Priors}}

{To evaluate the robustness of our method to noisy or low-quality priors, we conducted comprehensive experiments with controlled noise injection across different prior types. Following Pow3R~\citep{jang2025pow3r}, we designed realistic noise patterns that simulate real-world sensor inaccuracies and calibration errors.}

{\textbf{Noise injection settings.} For Camera Pose, we follow Pow3R and apply rotational noise by rotating the rotation matrix clockwise by 0° (clean), 20°, and 40°. In addition to rotation errors, we also consider isotropic Gaussian noise on the camera translation component with standard deviation $\sigma=0.0$ (clean), $0.05$, and $0.1$. For Camera Intrinsics, we follow Pow3R and scale the intrinsic parameters by factors of $1.0\times$ (clean), $0.8\times$, and $0.6\times$. For Depth Prior, we apply pixel-wise multiplicative noise to depth maps by scaling each pixel with a random factor $\sim \mathcal{N}(1.0, \sigma)$, where $\sigma=0.0$ (clean), $0.05$, and $0.1$.} {We evaluate reconstruction quality on 7-Scenes and DTU datasets, comparing against the baseline model without any priors (None). The results are shown in Tab.~\ref{tab:noise_robustness}.}

{Our model exhibits graceful performance degradation as noise increases, indicating robust feature learning. Even with moderate noise (20° rotation, $0.8\times$ intrinsics, $\sigma=0.05$ depth), noisy priors still provide meaningful guidance compared to the no-prior baseline in many cases. Camera pose priors show strong robustness up to 20° rotation error, maintaining better performance than the baseline. Depth priors are more sensitive to noise due to direct geometric guidance, but still maintain reasonable performance with $\sigma=0.05$. These results demonstrate that \modelname~can effectively leverage imperfect priors while maintaining robustness to various noise levels.}

\section{{More Analysis of Depth Prior Design}}
\subsection{{Depth Prior Injection: Addition vs. Concatenation}}
{We compared two depth prior injection strategies to validate our design choice: (1) Concat: Concatenate depth features with image tokens along the token dimension; (2) Addition (Ours): Directly add depth features to image tokens. The quantitative comparison is presented in Tab.~\ref{tab:depth_injection}.}

\begin{table*}[h]
\centering
\caption{\textbf{Comparison of depth prior injection strategies on 7-Scenes and DTU datasets.} Inference time and FLOPs are measured for depth prior injection on a single image with resolution $518 \times 378$.}
\label{tab:depth_injection}
\resizebox{\textwidth}{!}{
\begin{tabular}{l|cc|cc|cc}
\toprule
{\textbf{Depth Input}} & {\textbf{7S-Acc.} $\downarrow$} & {\textbf{7S-Comp.} $\downarrow$} & {\textbf{DTU-Acc.} $\downarrow$} & {\textbf{DTU-Comp.} $\downarrow$} & {\textbf{Inference Time (s)} $\downarrow$} & {\textbf{TFLOPs} $\downarrow$} \\
\midrule
{Concat} & {0.06} & {0.07} & {0.82} & {0.98} & {0.11} & {1.74} \\
{Addition (Ours)} & {\textbf{0.06}} & {\textbf{0.06}} & {\textbf{0.79}} & {\textbf{0.97}} & {\textbf{0.09}} & {\textbf{1.14}} \\
\bottomrule
\end{tabular}
}
\end{table*}

{Addition achieves superior accuracy with optimal efficiency, while concatenation increases computational cost by 52.6\% with comparable performance. Token addition offers two advantages: (1) Enhanced Spatial Alignment: Direct addition maintains pixel-wise correspondence between depth and image features, enabling more effective feature integration. (2) Computational Efficiency: Additive fusion introduces minimal overhead, avoiding expensive attention operations that would increase computational costs.}

\subsection{{Depth Normalization and Scale Information}}
{To demonstrate that depth prior normalization improves relative geometry, we conduct an ablation study comparing normalized vs. unnormalized depth inputs on 7-Scenes and DTU datasets, as shown in Tab.~\ref{tab:depth_normalization}.}

\begin{table*}[h]
\centering
\caption{\textbf{Comparison of normalized vs. unnormalized depth inputs on 7-Scenes and DTU datasets.} Normalized depth provides more stable geometric priors for relative-scale reconstruction.}
\label{tab:depth_normalization}
\resizebox{\textwidth}{!}{
\begin{tabular}{l|cccc|cccc}
\toprule
{\textbf{Depth Input}} & {\textbf{7S-Acc.} $\downarrow$} & {\textbf{7S-Comp.} $\downarrow$} & {\textbf{7S-NC1} $\uparrow$} & {\textbf{7S-NC2} $\uparrow$} & {\textbf{DTU-Acc.} $\downarrow$} & {\textbf{DTU-Comp.} $\downarrow$} & {\textbf{DTU-NC1} $\uparrow$} & {\textbf{DTU-NC2} $\uparrow$} \\
\midrule
{Unnormalized} & {0.11} & {0.12} & {0.66} & {0.67} & {2.60} & {5.96} & {0.52} & {0.51} \\
{Normalized (Ours)} & {\textbf{0.06}} & {\textbf{0.06}} & {\textbf{0.77}} & {\textbf{0.78}} & {\textbf{0.79}} & {\textbf{0.97}} & {\textbf{0.70}} & {\textbf{0.71}} \\
\bottomrule
\end{tabular}
}
\end{table*}

{The substantial improvement demonstrates that normalized depth effectively provides geometric priors for better relative-scale reconstruction. We hypothesize this is because normalization offers several key advantages: (1) Consistent Feature Range: Normalization maps depth values from diverse scenes with varying absolute scales into a unified [0,1] range, enabling the model to learn consistent depth-to-geometry mappings across different environments. (2) Improved Training Stability: Unnormalized depth values can vary by orders of magnitude across scenes, leading to unstable gradients during training. Normalization mitigates this issue by providing a bounded input space.}

\section{{More Details of Training and Inference}}

\subsection{{GPU Memory Requirements for Multi-View Processing}}
{We evaluate the GPU memory requirements on a single H20 GPU with input resolution of $518 \times 378$. The results are shown in Tab.~\ref{tab:gpu_memory}.}

\begin{table*}[h]
\centering
\caption{\textbf{GPU memory consumption (GB) for different prediction tasks across varying numbers of input views.} Measurements are performed on a single H20 GPU with resolution $518 \times 378$.}
\label{tab:gpu_memory}
\resizebox{0.55\textwidth}{!}{
\begin{tabular}{c|ccccc}
\toprule
{\textbf{N Views}} & {\textbf{Camera}} & {\textbf{Pointmap}} & {\textbf{Depth}} & {\textbf{Normal}} & {\textbf{3DGS}} \\
\midrule
{1} & {5.441} & {4.759} & {4.759} & {4.759} & {4.761} \\
{4} & {5.658} & {4.976} & {4.976} & {4.976} & {4.989} \\
{16} & {6.512} & {5.988} & {5.976} & {5.988} & {7.588} \\
{64} & {9.960} & {9.277} & {9.277} & {9.277} & {17.816} \\
{256} & {23.732} & {23.050} & {23.050} & {23.050} & {60.540} \\
{512} & {42.102} & {41.421} & {41.421} & {41.421} & {OOM} \\
\bottomrule
\end{tabular}
}
\end{table*}

{The memory consumption scales approximately linearly with the number of input views for most prediction tasks (camera, pointmap, depth, normal), growing from $\sim$5GB for single-view to $\sim$23GB for 256 views. However, the 3D Gaussian Splatting task exhibits significantly higher memory usage (60.5GB for 256 views), primarily due to additional convolutional layers in the GS head that decode high-dimensional Gaussian attributes (position, rotation, scaling, opacity, spherical harmonics) from dense feature maps. Further memory reduction could be achieved through optimized attention mechanisms and more compact 3D representations, following recent approaches like FastVGGT~\citep{shen2025fastvggt} and ReSplat~\citep{xu2025resplat}, which we leave for future work.}

\subsection{{Maximum Number of Input Views}}
{We evaluate the maximum number of input views that our model can handle on a single H20 GPU with input resolution of $518 \times 378$. The results are shown in Tab~\ref{tab:max_views}.}

\begin{table*}[h]
\centering
\tiny
\caption{\textbf{Maximum number of input views supported by different prediction tasks on a single H20 GPU with resolution $518 \times 378$.}}
\label{tab:max_views}
\resizebox{0.6\textwidth}{!}{
\begin{tabular}{lccccc}
\toprule
{\textbf{Task}} & {\textbf{Camera}} & {\textbf{Pointmap}} & {\textbf{Depth}} & {\textbf{Normal}} & {\textbf{3DGS}} \\
\midrule
{\textbf{Max Views}} & {1024} & {1024} & {1024} & {1024} & {360} \\
\bottomrule
\end{tabular}
}
\end{table*}

{Most dense prediction tasks (camera, pointmap, depth, normal) can handle up to 1024 input views, while 3D Gaussian Splatting is limited to 360 views due to its higher memory footprint from decoding high-dimensional Gaussian attributes.}

\subsection{{Training Time (Wall-Clock)}}
{We provide the wall-clock training time for full transparency and reproducibility. Our training was conducted on 32 NVIDIA H20 GPUs with gradient checkpointing for memory efficiency, Flash Attention v3 for accelerated attention computation, and mixed precision training (BF16).}

{The complete training process consists of two sequential stages. Stage 1 trains the dense prediction heads and requires approximately 42 hours of wall-clock time. Stage 2 trains the 3D Gaussian Splatting head and takes approximately 28 hours. The total end-to-end training time amounts to approximately 70 hours. Further acceleration could be achieved through advanced optimization techniques such as model parallelism (e.g., FSDP), FP8 precision training, or compiler-level optimizations (e.g., torch.compile).}

\subsection{Inference Speed and Practical Usability.}
We agree that a systematic comparison under uniform hardware is crucial. We provide median inference times on a single NVIDIA H20 GPU at a resolution of $518\times378$. The results are shown in Tab.~\ref{tab:inference_speed}.
\begin{table}[h]
\centering
\caption{\textbf{Inference speed comparison across different methods and view counts.} Median inference times are measured in milliseconds.}
\label{tab:inference_speed}
\begin{tabular}{lcccc}
\toprule
\textbf{Method} & \textbf{4 views} & \textbf{16 views} & \textbf{64 views} & \textbf{Tasks} \\
\midrule
VGGT & 218.1 & 967.3 & 7158.2 & Camera + Points + Depths \\
Pi3 & 210.2 & 908.6 & 6029.5 & Camera + Points + Depths \\
\modelname~ (Geo) & 218.3 & 963.4 & 7192.2 & Camera + Points + Depths \\
\modelname~ (Full) & 278.9 & 1196.2 & 8134.9 & All 5 Tasks \\
\bottomrule
\end{tabular}
\end{table}

As shown in Tab.~\ref{tab:inference_speed}, \modelname~ (Geo) matches VGGT's latency, indicating that our prior injection adds negligible computational cost. The full model introduces $\sim$13\% overhead at 64 views but completes all five tasks in a single forward pass. Regarding practical usability, the 3DGS head is modular and on-demand. At 64 views, the Geo model uses 9.96 GB of memory, while the Full model uses 17.82 GB; both fit comfortably on a single GPU. Furthermore, techniques such as chunked rendering or efficient attention mechanisms~\citep{deng2025vggt,shen2025fastvggt} can be employed to further reduce memory consumption for higher view counts.

\section{Limitations and Future Works}
Despite the promising results achieved by our approach, several limitations remain. First, our method demonstrates suboptimal performance on dynamic scenes and autonomous driving environments~\citep{miao2025advances}, primarily due to the under-representation of such data in our training distribution. We plan to address this through strategic dataset expansion to enhance model generalization~\citep{wang2025pi}. In addition, incorporating physical motion priors~\citep{quan2026particlegs} to enhance the model’s capability in dynamic environments represents a promising direction for future research. Additionally, our current implementation supports input resolutions ranging from 300 to 700 pixels and cannot effectively handle scenarios where the number of input views reaches into the thousands. This constraint becomes particularly apparent when running on consumer-grade GPUs. Future work will explore computational optimizations~\cite{sun2020test} to improve model efficiency and enable processing of longer visual sequences with reduced memory requirements.

\section{More Visual Results}
\subsection{Novel view synthesis}
\begin{figure}[!t]
    \centering
    \includegraphics[width=0.9\textwidth]{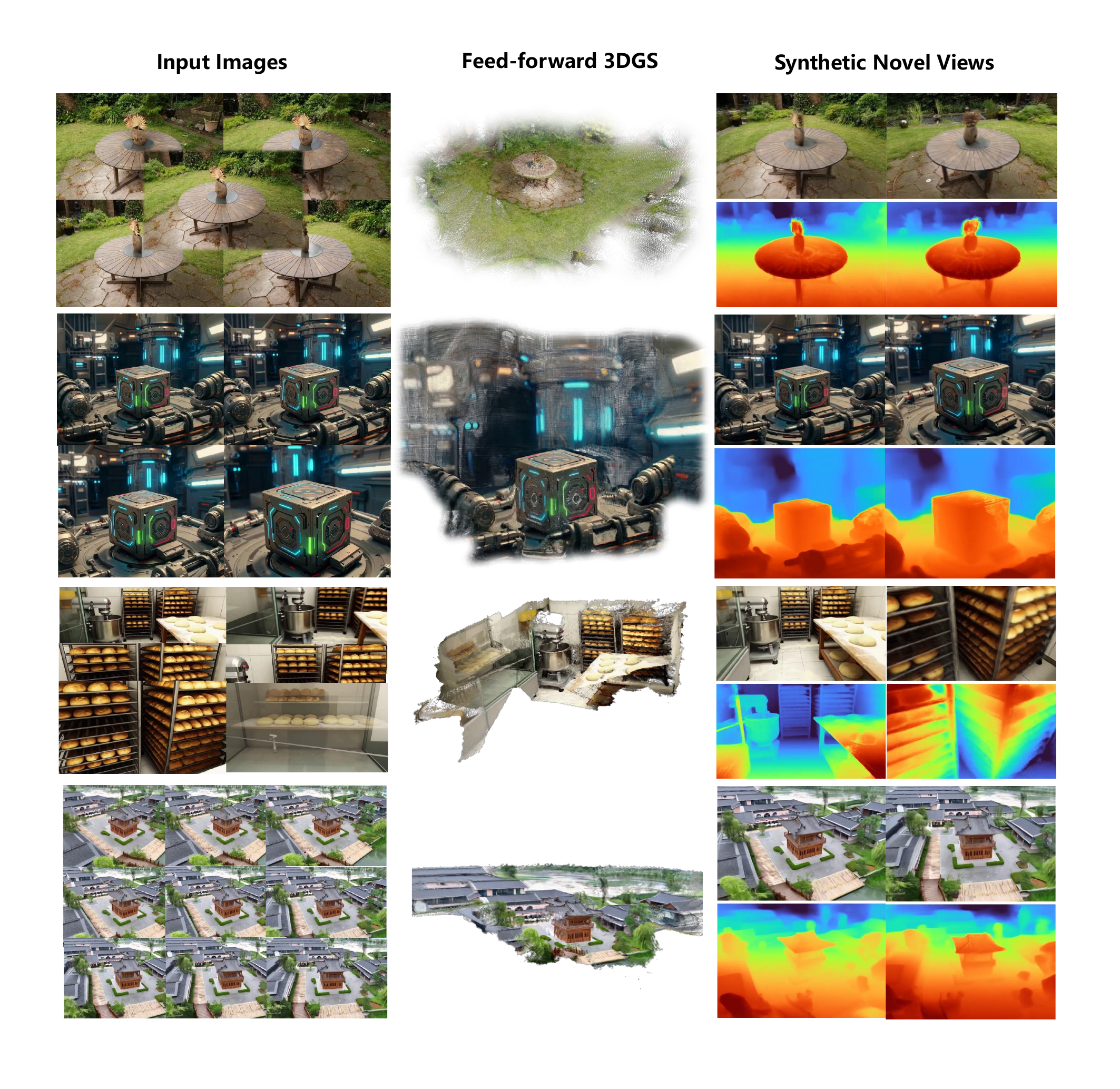}
    \caption{\textbf{Visual Results of Feed-Forward 3D Gaussians Generated by \modelname.}}
    \label{fig:ffgs_vis_add}
\end{figure}

In Fig.~\ref{fig:ffgs_vis_add}, we present additional results of feedforward Gaussians and their corresponding novel view renderings. Whether the input consists of AI-generated videos or real multi-view images, our method consistently infers 3D Gaussian splatting with plausible geometric structures and renders high-quality novel view images. This demonstrates that our model generalizes effectively across diverse input scenarios.

\subsection{Point map reconstruction}
\begin{figure}[!t]
    \centering
    \includegraphics[width=1.0\textwidth]{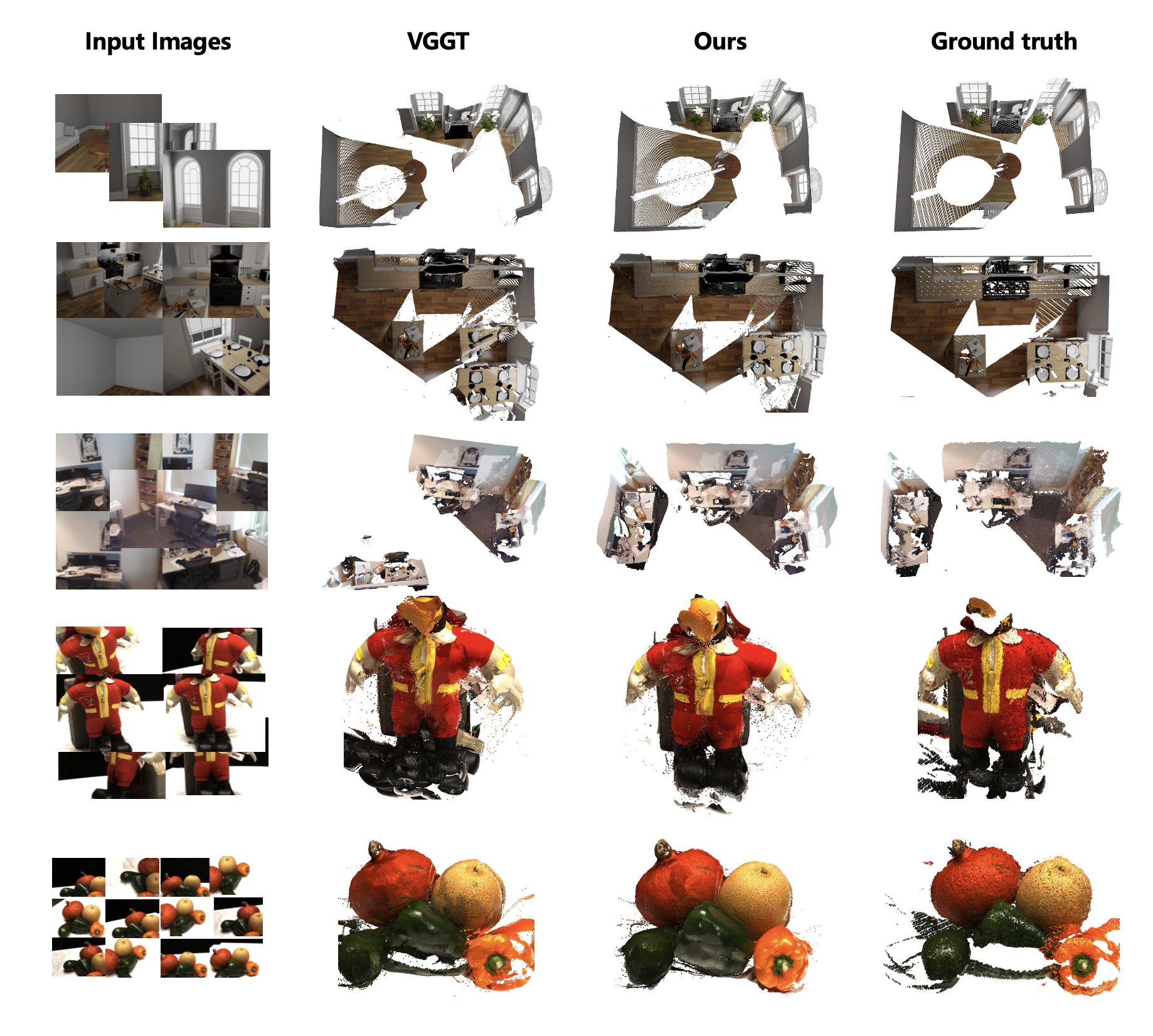}
    \caption{\textbf{Visual Comparisons on 7-Scenes, NRGBD, and DTU datasets.} \modelname~delivers superior reconstruction fidelity compared to VGGT, effectively capturing spatial relationships within scenes while producing geometrically coherent structures.
    }
    \label{fig:pointmap_wgt}
\end{figure}

\begin{figure}[!t]
    \centering
    \includegraphics[width=1.0\textwidth]{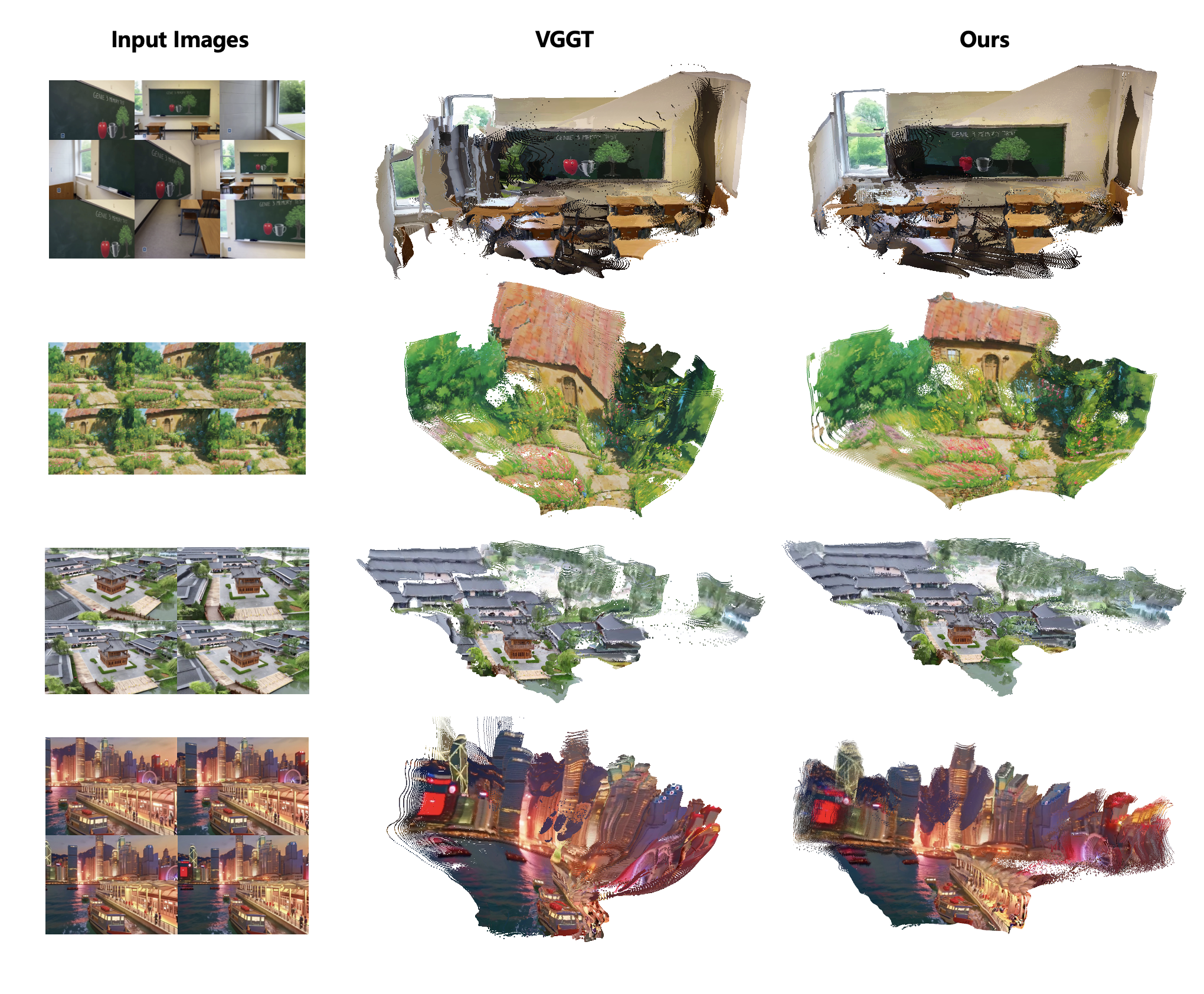}
    \caption{\textbf{Visual Comparisons of In-The-Wild Multi-View 3D Reconstruction.} \modelname~delivers superior reconstruction fidelity with in-the-wild images as input, generating more plausible results in challenging scenarios compared to VGGT. Our approach effectively resolves complex spatial arrangements and maintains geometric consistency even when confronted with difficult viewing conditions, occlusions, or intricate environmental structures.
    }
    \label{fig:pointmap_wogt}
\end{figure}

We provide additional visual comparisons of point map reconstruction in Fig.~\ref{fig:pointmap_wgt} and Fig.~\ref{fig:pointmap_wogt}. Fig.~\ref{fig:pointmap_wgt} features selected scenes from 7-scenes, NRGBD, and DTU datasets, where comparisons with ground truth reveal that \modelname~produces more consistent reconstructions, particularly when processing sparse viewpoints that require inference of spatial distributions. In Fig.~\ref{fig:pointmap_wogt}, we evaluate model performance on in-the-wild images by processing both video generation model outputs and real-world multi-view captures. The results demonstrate that \modelname~generates geometrically coherent and plausible reconstructions across these diverse inputs, highlighting its strong generalization capabilities.

\section{Other Applications}
\subsection{Surface Reconstruction.}
\begin{figure}[!t]
    \centering
    \includegraphics[width=1.0\textwidth]{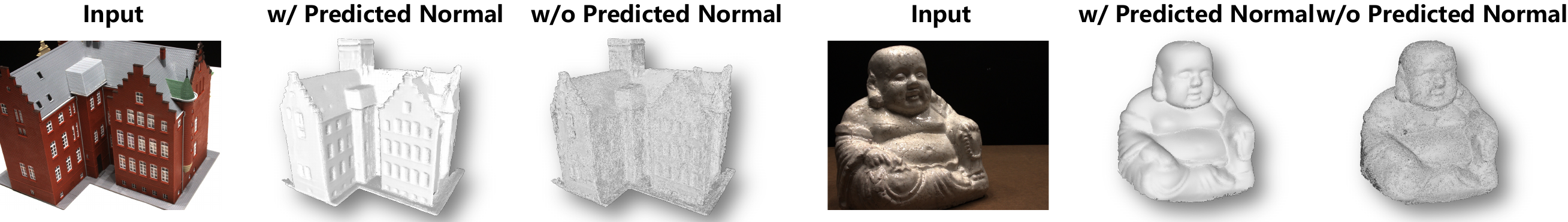}
    \caption{\textbf{\modelname~ Improves Surface Reconstruction with Predicted Normal Maps.}
    }
    \label{fig:surface_reconstruction}
\end{figure}

\modelname~supports high-quality 3D surface reconstruction with the predicted smooth normal maps. As shown in~\reffig{surface_reconstruction}, by leveraging the predicted normals instead of traditional geometric normal estimation from point clouds, \modelname~ produces a cleaner surface with sharp details via Poisson surface reconstruction~\citep{kazhdan2006poisson}.

\end{document}

%% file: math_commands.tex
\usepackage{amsmath,amsfonts,bm}

\def\eqref#1{equation~\ref{#1}}

\def\1{\bm{1}}

\def\vc{{\bm{c}}}

\def\vq{{\bm{q}}}

\def\vt{{\bm{t}}}

\def\mD{{\bm{D}}}
\def\mE{{\bm{E}}}
\def\mF{{\bm{F}}}

\def\mI{{\bm{I}}}

\def\mK{{\bm{K}}}

\def\mN{{\bm{N}}}

\def\mP{{\bm{P}}}

\def\mR{{\bm{R}}}

\def\mT{{\bm{T}}}

\DeclareMathAlphabet{\mathsfit}{\encodingdefault}{\sfdefault}{m}{sl}
\SetMathAlphabet{\mathsfit}{bold}{\encodingdefault}{\sfdefault}{bx}{n}

\newcommand{\Ls}{\mathcal{L}}
\newcommand{\R}{\mathbb{R}}



%% file: Tables/pointmap.tex
\begin{table*}[t]
    \centering
    \caption{
        \textbf{Point map Reconstruction on 7-Scenes, NRGBD, and DTU.} We report the performance of WorldMirror under different input configurations. \colorbox{bestblue}{Best} and \colorbox{secondblue}{second best} results are highlighted.
    }
    \resizebox{1.0\textwidth}!{
    \begin{tabular}{lcccccccccccc}
        \toprule[0.14em]
        {\multirow{4}{*}{\textbf{Method}}} &
        \multicolumn{4}{c}{\textbf{7-Scenes} (scene)} &
        \multicolumn{4}{c}{\textbf{NRGBD} (scene)} & 
        \multicolumn{4}{c}{\textbf{DTU} (object)} \\
        \cmidrule(r){2-5} \cmidrule(r){6-9} \cmidrule(r){10-13}
        & \multicolumn{2}{c}{Acc. $\downarrow$} & \multicolumn{2}{c}{Comp. $\downarrow$} 
        & \multicolumn{2}{c}{Acc. $\downarrow$} & \multicolumn{2}{c}{Comp. $\downarrow$} 
        & \multicolumn{2}{c}{Acc. $\downarrow$} & \multicolumn{2}{c}{Comp. $\downarrow$} \\
        \cmidrule(r){2-3} \cmidrule(r){4-5} \cmidrule(r){6-7} \cmidrule(r){8-9} \cmidrule(r){10-11} \cmidrule(r){12-13}
        & Mean & Med. & Mean & Med. & Mean & Med. & Mean & Med. & Mean & Med. & Mean & Med. \\
        \midrule[0.08em]
        Fast3R~\citep{yang2025fast3r} & 0.096 & 0.065 & 0.145 & 0.093 & 0.135 & 0.091 & 0.163 & 0.104 & 3.340 & 1.919 & 2.929 & 1.125 \\
        CUT3R~\citep{wang2025continuous} & 0.094 & 0.051 & 0.101 & 0.050 & 0.104 & 0.041 & 0.079 & 0.031 & 4.742 & 2.600 & 3.400 & 1.316 \\
        FLARE~\citep{zhang2025flare} & 0.085 & 0.058 & 0.142 & 0.104 & 0.053 & 0.024 & 0.051 & 0.025 & 2.541 & 1.468 & 3.174 & 1.420 \\
        VGGT~\citep{wang2025vggt} & 0.046 & 0.026 & 0.057 & 0.034 & 0.051 & 0.029 & 0.066 & 0.038 & 1.338 & 0.779 & 1.896 & 0.992 \\
        $\pi^3$\citep{wang2025pi} & 0.048 & 0.028 & 0.072 & 0.047 & \cellcolor{secondblue}0.026 & \cellcolor{secondblue}0.015 & \cellcolor{secondblue}0.028 & \cellcolor{secondblue}0.014 & 1.198 & 0.646 & 1.849 & 0.607 \\
        \midrule[0.08em]
        WorldMirror & 0.043 & 0.026 & 0.049 & 0.028 & 0.041 & 0.020 & 0.045 & 0.019 & 1.017 & 0.564 & 1.780 & 0.690 \\
        WorldMirror (w/ intrinsics) & 0.042 & 0.028 & 0.048 & 0.026 & 0.041 & 0.020 & 0.045 & 0.019 & 0.977 & 0.542 & 1.762 & 0.682 \\
        WorldMirror (w/ depth) & 0.038 & 0.024 & 0.039 & 0.023 & 0.032 & 0.015 & 0.031 & 0.014 & \cellcolor{secondblue}0.831 & \cellcolor{secondblue}0.506 & \cellcolor{secondblue}1.022 & \cellcolor{secondblue}0.599 \\
        WorldMirror (w/ camera pose) & \cellcolor{secondblue}0.023 & \cellcolor{secondblue}0.014 & \cellcolor{secondblue}0.036 & \cellcolor{secondblue}0.019 & 0.029 & 0.018 & 0.032 & 0.017 & 0.990 & 0.548 & 1.847 & 0.686 \\
        WorldMirror (w/ intrinsics/depth/camera pose) & \cellcolor{bestblue}0.018 & \cellcolor{bestblue}0.011 & \cellcolor{bestblue}0.023 & \cellcolor{bestblue}0.014 & \cellcolor{bestblue}0.016 & \cellcolor{bestblue}0.011 & \cellcolor{bestblue}0.014 & \cellcolor{bestblue}0.010 & \cellcolor{bestblue}0.735 & \cellcolor{bestblue}0.461 & \cellcolor{bestblue}0.935 & \cellcolor{bestblue}0.550 \\
        \bottomrule[0.14em]
    \end{tabular}
    }
    \label{tab:pointmap}
\end{table*}

%% file: Tables/camera.tex
\begin{table*}[!t]
\centering
\caption{\textbf{Camera Pose Estimation on RealEstate10K, Sintel, and TUM-dynamics.} All datasets are excluded from the training set, except that RealEstate10K was included for CUT3R training. \colorbox{bestblue}{Best} and \colorbox{secondblue}{second best} results are highlighted.}
\resizebox{0.9\textwidth}{!}{
\begin{tabular}{lccccccccc}
\toprule[0.14em]
\multicolumn{1}{l}{\multirow{3}{*}{\textbf{Method}}} &
\multicolumn{3}{c}{\textbf{RealEstate10K} (mixed, static)} &
\multicolumn{3}{c}{\textbf{Sintel} (outdoor, dynamic)} &
\multicolumn{3}{c}{\textbf{TUM-dynamics} (indoor, dynamic)} \\
\cmidrule(r){2-4} \cmidrule(r){5-7} \cmidrule(r){8-10}
\multicolumn{1}{c}{} &
RRA@30 $\uparrow$ & RTA@30 $\uparrow$ & AUC@30 $\uparrow$ &
ATE$\downarrow$ & RPE trans$\downarrow$ & RPE rot$\downarrow$ &
ATE$\downarrow$ & RPE trans$\downarrow$ & RPE rot$\downarrow$ \\
\midrule[0.08em]
Fast3R\citep{yang2025fast3r} & 99.05 & 81.86 & 61.68 & 0.371 & 0.298 & 13.75 & 0.090 & 0.101 & 1.425 \\
CUT3R~\citep{wang2025continuous} & 99.82 & 95.10 & 81.47 & 0.217 & 0.070 & 0.636 & 0.047 & 0.015 & 0.451 \\
FLARE~\citep{zhang2025flare} & 99.69 & 95.23 & 80.01 & 0.207 & 0.090 & 3.015 & 0.026 & 0.013 & 0.475 \\
VGGT~\citep{wang2025vggt} & \cellcolor{secondblue}99.97 & 93.13 & 77.62 & 0.167 & 0.062 & 0.491 & \cellcolor{secondblue}0.012 & \cellcolor{secondblue}0.010 & \cellcolor{secondblue}0.312 \\
$\pi^3$~\citep{wang2025pi} & \cellcolor{bestblue}99.99 & \cellcolor{secondblue}95.62 & \cellcolor{secondblue}85.90 & \cellcolor{bestblue}0.074 & \cellcolor{bestblue}0.040 & \cellcolor{bestblue}0.282 & 0.014 & \cellcolor{bestblue}0.009 & \cellcolor{secondblue}0.312 \\
\midrule
WorldMirror & \cellcolor{bestblue}99.99 & \cellcolor{bestblue}95.81 & \cellcolor{bestblue}86.28 & \cellcolor{secondblue}0.096 & \cellcolor{secondblue}0.058 & \cellcolor{secondblue}0.490 & \cellcolor{bestblue}0.010 & \cellcolor{bestblue}0.009 & \cellcolor{bestblue}0.297 \\
\bottomrule[0.14em]
\end{tabular}
}
\label{tab:camera_pose}
\end{table*}

%% file: Tables/normal.tex
\begin{table*}[t]
\centering
\caption{\textbf{Surface Normal Estimation on ScanNet, NYUv2, and iBims-1.} We compare with both regression-based and diffusion-based surface normal estimation approaches. \colorbox{bestblue}{Best} and \colorbox{secondblue}{second best} results are highlighted.}
\resizebox{1.0\textwidth}{!}{
    \begin{tabular}{lcccccccccccc}
    \toprule[0.14em]
    \multicolumn{1}{l}{\multirow{3}{*}{\textbf{Method}}} &
    \multicolumn{4}{c}{\textbf{ScanNet}} &
    \multicolumn{4}{c}{\textbf{NYUv2}} &
    \multicolumn{4}{c}{\textbf{iBims-1}} \\
    \cmidrule(r){2-5} \cmidrule(r){6-9} \cmidrule(r){10-13}
    \multicolumn{1}{c}{} &
    mean $\downarrow$ & med $\downarrow$ & $22.5^\circ\uparrow$ & $30^\circ\uparrow$ &
    mean $\downarrow$ & med $\downarrow$ & $22.5^\circ\uparrow$ & $30^\circ\uparrow$ &
    mean $\downarrow$ & med $\downarrow$ & $22.5^\circ\uparrow$ & $30^\circ\uparrow$ \\
    \midrule[0.08em]
    OASIS~\citep{chen2020oasis} & 32.8 & 28.5 & 38.5 & 52.6 & 29.2 & 23.4 & 48.4 & 60.7 & 32.6 & 24.6 & 46.6 & 57.4 \\
    EESNU~\citep{bae2021estimating} & - & - & - & - & 16.2 & 8.5 & 77.2 & 83.5 & 20.0 & 8.4 & 73.4 & 78.2 \\
    Omnidata v1~\citep{eftekhar2021omnidata} & 22.9 & 12.3 & 66.1 & 73.2 & 23.1 & 12.9 & 66.3 & 73.6 & 19.0 & 7.5 & 76.1 & 80.1 \\
    Omnidata v2~\citep{kar20223d} & 16.2 & 8.5 & 79.5 & 84.7 & 17.2 & 9.7 & 76.5 & 83.0 & 18.2 & 7.0 & 77.4 & 81.1 \\
    DSine~\citep{bae2024rethinking} & 16.2 & \cellcolor{secondblue}8.3 & 78.7 & 84.4 & \cellcolor{secondblue}16.4 & \cellcolor{secondblue}8.4 & 77.7 & 83.5 & \cellcolor{secondblue}17.1 & \cellcolor{bestblue}6.1 & 79.0 & 82.3 \\
    GeoWizard~\citep{fu2024geowizard} & 16.7 & 9.5 & 78.3 & 84.2 & 19.5 & 11.7 & 74.5 & 81.6 & 20.4 & 9.4 & 76.4 & 80.6 \\
    StableNormal~\citep{ye2024stablenormal} & \cellcolor{secondblue}16.0 & 9.9 & \cellcolor{secondblue}81.5 & \cellcolor{secondblue}86.5 & 18.5 & 11.2 & \cellcolor{secondblue}77.5 & \cellcolor{secondblue}83.6 & 17.9 & 8.5 & \cellcolor{bestblue}80.4 & \cellcolor{bestblue}83.9 \\
    \midrule
    WorldMirror & \cellcolor{bestblue}13.8 & \cellcolor{bestblue}7.3 & \cellcolor{bestblue}82.5 & \cellcolor{bestblue}87.3 & \cellcolor{bestblue}15.1 & \cellcolor{bestblue}8.0 & \cellcolor{bestblue}80.1 & \cellcolor{bestblue}85.7 & \cellcolor{bestblue}16.6 & \cellcolor{secondblue}6.4 & \cellcolor{secondblue}80.1 & \cellcolor{secondblue}83.7 \\
    \bottomrule[0.14em]
    \end{tabular}
}
\label{tab:normal_estimation}
\end{table*}

%% file: Tables/nvs_multi_resolution.tex
\begin{table*}[t]
\centering
\caption{\textbf{Multi-resolution novel view synthesis evaluation on DL3DV. } $*$ denotes using the pose-free optimization~\citep{ye2024no} for fair comparison with non–pose-free baselines~\citep{xu2025depthsplat}. \colorbox{bestblue}{Best} results for each type are highlighted.}
\label{tab:dl3dv_multiresolution_depthsplat}

\resizebox{1.0\textwidth}{!}{
\begin{tabular}{lccccccccccc}
\toprule[0.14em]
\multicolumn{1}{l}{\multirow{3}{*}{{\textbf{Method}}}} &
\multicolumn{1}{l}{\multirow{3}{*}{{\textbf{Prior Condition}}}} &
\multicolumn{1}{l}{\multirow{3}{*}{{\textbf{Resolution}}}} &
\multicolumn{3}{c}{{\textbf{DL3DV} (8 Views)}} &
\multicolumn{3}{c}{{\textbf{DL3DV} (24 Views)}} &
\multicolumn{3}{c}{{\textbf{DL3DV} (64 Views)}} \\
\cmidrule(r){4-6} \cmidrule(r){7-9} \cmidrule(r){10-12}
& & & 
{PSNR $\uparrow$} &
{SSIM $\uparrow$} &
{LPIPS $\downarrow$} &
{PSNR $\uparrow$} &
{SSIM $\uparrow$} &
{LPIPS $\downarrow$} &
{PSNR $\uparrow$} &
{SSIM $\uparrow$} &
{LPIPS $\downarrow$} \\
\midrule

{AnySplat} & {None}
& {448$\times$252} 
& {15.62} & {0.453} & {0.305} 
& {18.68} & {0.571} & \cellcolor{bestblue}{0.221} 
& {19.50} & {0.610} & \cellcolor{bestblue}{0.208} \\

{WorldMirror} & {None}
& {448$\times$252} 
& \cellcolor{bestblue}{17.50} & \cellcolor{bestblue}{0.518} & \cellcolor{bestblue}{0.303} 
& \cellcolor{bestblue}{19.15} & \cellcolor{bestblue}{0.602} & {0.241} 
& \cellcolor{bestblue}{19.51} & \cellcolor{bestblue}{0.626} & {0.239} \\
\midrule

{FLARE} & {Intrinsics}
& {256$\times$256} 
& {14.77} & {0.412} & {0.647} 
& {14.11} & {0.398} & {0.761} 
& {-} & {-} & {-} \\

{WorldMirror} & {Intrinsics}
& {252$\times$252} 
& \cellcolor{bestblue}{16.83} & \cellcolor{bestblue}{0.480} & \cellcolor{bestblue}{0.320} 
& \cellcolor{bestblue}{18.76} & \cellcolor{bestblue}{0.574} & \cellcolor{bestblue}{0.244} 
& {19.10} & {0.593} & {0.240} \\
\midrule

{DepthSplat} & {Intrinsics + Poses}
& {448$\times$256} 
& {18.79} & {0.619} & {0.316} 
& {18.71} & {0.643} & {0.313} 
& {16.80} & {0.551} & {0.416} \\

{$\text{WorldMirror}^{*}$} & {Intrinsics + Poses}
& {448$\times$252} 
& \cellcolor{bestblue}{19.08} & \cellcolor{bestblue}{0.624} & \cellcolor{bestblue}{0.261} 
& \cellcolor{bestblue}{20.24} & \cellcolor{bestblue}{0.675} & \cellcolor{bestblue}{0.221} 
& \cellcolor{bestblue}{20.30} & \cellcolor{bestblue}{0.680} & \cellcolor{bestblue}{0.226} \\

\bottomrule[0.14em]
\end{tabular}
}
\end{table*}

%% file: Figs/NVS_result1.tex
\begin{figure*}[!t]
    \centering

    \includegraphics[width=0.99\textwidth]{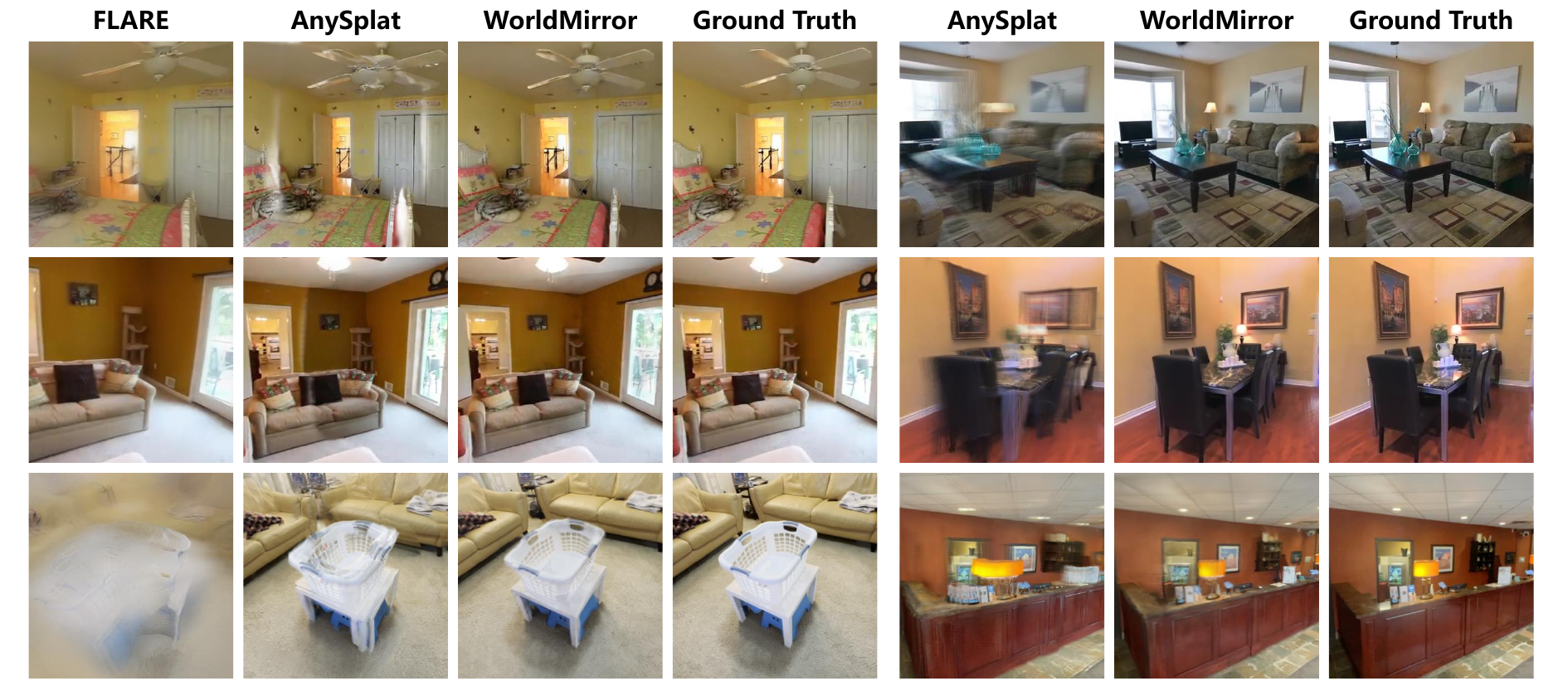}

    \caption{\textbf{Qualitative Comparisons of Novel View Synthesis.} We compare with FLARE and AnySplat on RealEstate10K and DL3DV. The first four columns correspond to the sparse-view setting, while the latter three correspond to the dense-view setting. Our approach surpasses baselines in both appearance fidelity and geometric perception.
}
    \label{fig:nvs_result1}
\end{figure*}

%% file: Tables/ablation_prior.tex
\begin{table*}[t]
\centering
\caption{\textbf{Prior Embedding Ablation.} Results are averaged over ETH3D and DTU datasets with 10 views as input. `Single token' offers both superior performance and high efficiency. \colorbox{bestblue}{Best} results in each group are highlighted.}
\label{tab:ablation-prior}
\resizebox{1.0\textwidth}{!}{%
\begin{tabular}{llccccccccc}
\toprule[0.14em]
\multirow{2}{*}{\textbf{Inputs}} & \multirow{2}{*}{\textbf{Prior Embedding}} & \textbf{Extra} & \textbf{Focal} & \textbf{Depth} & \multicolumn{3}{c}{\textbf{Pose}} & \textbf{Point} & \multirow{2}{*}{\textbf{Avg.} $\uparrow$} \\
 & & \textbf{Params} & acc@1.03$\uparrow$ & $\tau$@1.03 $\uparrow$ & RRA@5 $\uparrow$ & RTA@5 $\uparrow$ & AUC@5 $\uparrow$ & $\tau$@1.03 $\uparrow$ & \\
\midrule[0.08em]
\multirow{2}{*}{Images \& Poses} & Dense Plücker & 9.02M & 33.07 & \cellcolor{bestblue}31.00 & 98.59 & \cellcolor{bestblue}93.52 & 72.74 & 33.74 & 60.44 \\
 & Single Token & \cellcolor{bestblue}1.06M & \cellcolor{bestblue}33.82 & 28.02 & \cellcolor{bestblue}98.89 & 92.57 & \cellcolor{bestblue}74.55 & \cellcolor{bestblue}38.51 & \cellcolor{bestblue}61.06 \\
\midrule[0.08em]
\multirow{2}{*}{Images \& Intrinsics} & Dense Raymap & 6.65M & \cellcolor{bestblue}86.48 & 29.36 & 97.17 & 88.48 & 60.57 & \cellcolor{bestblue}37.40 & 66.58 \\
 & Single Token & \cellcolor{bestblue}1.06M & 84.43 & \cellcolor{bestblue}34.70 & \cellcolor{bestblue}98.18 & \cellcolor{bestblue}93.64 & \cellcolor{bestblue}66.52 & 36.29 & \cellcolor{bestblue}68.96 \\
\bottomrule[0.14em]
\end{tabular}
}
\end{table*}

%% file: Tables/abaltion_gs.tex
\begin{table*}[!tbp]
\centering
\caption{\textbf{Novel View Synthesis Ablation.} \colorbox{bestblue}{Best} and \colorbox{secondblue}{second best} results are highlighted.}
\label{tab:ablation-gs}
\resizebox{0.9\textwidth}{!}{%
\begin{tabular}{lccccccccc}
\toprule[0.14em]
\multirow{2}{*}{\textbf{Method}} & \multicolumn{3}{c}{\textbf{RealEstate10K} (2 views)} & \multicolumn{3}{c}{\textbf{DL3DV} (8 views)} & \multicolumn{3}{c}{\textbf{VR-NeRF} (32 views)} \\ \cmidrule(r){2-4} \cmidrule(r){5-7} \cmidrule(r){8-10} 
 & PSNR $\uparrow$ & SSIM $\uparrow$ & LPIPS $\downarrow$ & PSNR $\uparrow$ & SSIM $\uparrow$ & LPIPS $\downarrow$ & PSNR $\uparrow$ & SSIM $\uparrow$ & LPIPS $\downarrow$ \\
\midrule[0.08em]
w/o GT Cameras & \cellcolor{bestblue}20.30 & \cellcolor{secondblue}0.691 & \cellcolor{secondblue}0.193 & \cellcolor{secondblue}20.69 & 0.666 & 0.206 & 24.76 & 0.788 & \cellcolor{secondblue}0.197 \\
w/o Novel Views & 18.51 & 0.651 & 0.215 & 20.21 & 0.664 & \cellcolor{bestblue}0.196 & 24.35 & 0.781 & 0.199 \\
w/o GS DPT & 20.28 & \cellcolor{bestblue}0.693 & 0.195 & 20.55 & \cellcolor{secondblue}0.667 & 0.218 & \cellcolor{secondblue}25.08 & \cellcolor{secondblue}0.798 & \cellcolor{bestblue}0.191 \\
\midrule[0.08em]
Ours & \cellcolor{secondblue}20.29 & \cellcolor{bestblue}0.693 & \cellcolor{bestblue}0.192 & \cellcolor{bestblue}20.91 & \cellcolor{bestblue}0.671 & \cellcolor{secondblue}0.198 & \cellcolor{bestblue}25.75 & \cellcolor{bestblue}0.811 & 0.198 \\
\bottomrule[0.14em]
\end{tabular}
}
\end{table*}

%% file: Tables/depth.tex
\begin{table*}[t]
\centering
\caption{\textbf{Monocular and Video Depth Estimation on NYUv2, Sintel, and KITTI.}}
\resizebox{1.0\textwidth}{!}{
    \begin{tabular}{lcccccccc}
    \toprule[0.14em]
\multirow{3}{*}{\textbf{Method}} & \multicolumn{2}{c}{\textbf{NYU-v2} (Monocular)} & \multicolumn{2}{c}{\textbf{Sintel} (Monocular)} & \multicolumn{2}{c}{\textbf{KITTI} (Video)} & \multicolumn{2}{c}{\textbf{Sintel} (Video)} \\ \cmidrule(r){2-3} \cmidrule(r){4-5} \cmidrule(r){6-7} \cmidrule(r){8-9}
                        & Abs Rel $\downarrow$          & $\delta < 1.25 \uparrow$          & Abs Rel $\downarrow$          & $\delta < 1.25 \uparrow$         & Abs Rel $\downarrow$        & $\delta < 1.25 \uparrow$       & Abs Rel  $\downarrow$       & $\delta < 1.25 \uparrow$        \\ \midrule
DUSt3R~\citep{wang2024dust3r}                  & 0.081             & 0.909              & 0.488             & 0.532              & 0.143           & 0.814           & 0.662           & 0.434            \\
MASt3R~\citep{leroy2024grounding}                  & 0.11              & 0.865              & 0.413             & 0.569              & 0.115           & 0.848           & 0.558           & 0.487            \\
MonST3R~\citep{zhang2024monst3r}                 & 0.094             & 0.887              & 0.492             & 0.525              & 0.107           & 0.884           & 0.399           & 0.519            \\
Fast3R~\citep{yang2025fast3r}                  & 0.093             & 0.898              & 0.544             & 0.509              & 0.138           & 0.834           & 0.638           & 0.422            \\
CUT3R~\citep{wang2025continuous}                   & 0.081             & 0.914              & 0.418             & 0.52               & 0.122           & 0.876           & 0.417           & 0.507            \\
FLARE~\citep{zhang2025flare}                   & 0.089             & 0.898              & 0.606             & 0.402              & 0.356           & 0.57            & 0.729           & 0.336            \\
VGGT~\citep{wang2025vggt}                    & 0.056             & 0.951              & 0.606             & 0.599              & \underline{0.062}           & \underline{0.969}           & 0.299           & 0.638            \\
$\pi^3$~\citep{wang2025pi}                & \underline{0.054}             & \underline{0.956}              & \textbf{0.277}             & \underline{0.614}              & \textbf{0.038}           & \textbf{0.986}           & \textbf{0.233}          & \underline{0.664}            \\
\midrule
    \textbf{WorldMirror}                    & \textbf{0.052}             & \textbf{0.957}              & \underline{0.339}             & \textbf{0.624}              & 0.063           & 0.968           & \underline{0.289}           & \textbf{0.668}  \\
    \bottomrule[0.14em]
    \end{tabular}
}
\label{tab:depth_estimation}
\end{table*}

%% file: Tables/nvs2.tex
\begin{table*}[!t]
\centering
\caption{\textbf{Novel View Synthesis with 3DGS Optimization on RealEsate10K, DL3DV, and VRNeRF.} In Post-Optimization, the \textit{random point cloud} refers to initializing Gaussian positions randomly, whereas the \textit{predicted point cloud} uses the point cloud estimated by our method as the initialization of Gaussian positions.}
\resizebox{1.0\textwidth}{!}{

    \begin{tabular}{lccccccccccccc}
    \toprule[0.14em]
    \multicolumn{1}{l}{\multirow{3}{*}{\textbf{Method}}} & \multicolumn{1}{l}{\multirow{3}{*}{\textbf{Iterations}}} & \multicolumn{4}{c}{\textbf{RealEstate10K } (32 views)} & \multicolumn{4}{c}{\textbf{DL3DV} (64 views)} & \multicolumn{4}{c}{\textbf{VRNeRF} (64 views)} \\ \cmidrule(r){3-6} \cmidrule(r){7-10} \cmidrule(r){11-14}
     \multicolumn{1}{c}{}        &               & PSNR $\uparrow$        & SSIM $\uparrow$      & LPIPS $\downarrow$  & Time $\downarrow$     & PSNR $\uparrow$       & SSIM $\uparrow$      & LPIPS $\downarrow$   & Time $\downarrow$   & PSNR $\uparrow$        & SSIM $\uparrow$       & LPIPS $\downarrow$   & Time $\downarrow$   \\ \midrule[0.08em]
     \textbf{Feedforward} \\
    AnySplat      & -          & 19.96       & 0.718       & 0.234  &  $<$2s  & 18.40       & 0.602      & 0.286   & $<$2s  & 22.11       & 0.759      & 0.288  &  $<$2s  \\
    WorldMirror     & -               & 25.14        & 0.859      & 0.109  & $<$2s   & 21.25      & 0.703   & \underline{0.223}  & $<$2s   & 25.77       & 0.830      & \textbf{0.208}  & $<$2s   \\ \midrule[0.08em]
    \textbf{Post Optimization} \\
    \textit{random points cloud}     & 3,000             & 26.03       & 0.875      & 0.145  & 19s   & \underline{23.61}       & 0.765       & 0.244 & 21s    & \textbf{26.45}       & 0.840       & 0.259    & 21s  \\
    \textit{predicted points cloud}    & 1,000               & \underline{27.29}       & \underline{0.906}      & \underline{0.092}  &  10s  & 23.43       & \underline{0.772}       & 0.248 &  12s   & 25.19       & \underline{0.841}       & 0.257 & 11s      \\
    AnySplat   & 1,000       & 23.85            &  0.834          &  0.192   & 23s      &   20.84          &   0.695         &  0.287     &  55s   & 23.19            &  0.782          &  0.322     &  33s   \\
    AnySplat   & 3,000       &  26.03           &   0.870         & 0.155    & 56s      &  22.20           &  0.723          &  0.226     & 126s    & 24.64            &  0.798          &  0.272     &  65s   \\
    WorldMirror   & 1,000           & \textbf{27.79}       & \textbf{0.915}      & \textbf{0.076}   & 23s  & \textbf{23.86}        & \textbf{0.786}      & \textbf{0.172}  &  45s  & \underline{25.98}       & \textbf{0.845}      & \underline{0.214}   & 38s  \\ \bottomrule[0.14em]
    \end{tabular}

}

\label{tab:novel_view_synthesis2}
\end{table*}

%% file: Tables/nvs1.tex
\begin{table*}[!t]
\centering
\caption{\textbf{Novel View Synthesis on RealEstate10K and DL3DV.} We compare with feed-forward 3DGS methods under sparse and dense-view settings at the resolution of $518\times378$. FLARE focuses on sparse views NVS and thus its performance under dense-view settings is omitted.}
\resizebox{1.0\textwidth}{!}{
    \begin{tabular}{lcccccccccccc}
    \toprule[0.14em]
    \multicolumn{1}{l}{\multirow{3}{*}{\textbf{Method}}} & \multicolumn{3}{c}{\textbf{RealEstate10K} (2 views)} & \multicolumn{3}{c}{\textbf{DL3DV} (8 views)} & \multicolumn{3}{c}{\textbf{RealEstate10K} (32 views)} & \multicolumn{3}{c}{\textbf{DL3DV} (64 views)} \\ \cmidrule(r){2-4} \cmidrule(r){5-7} \cmidrule(r){8-10} \cmidrule(r){11-13}
     \multicolumn{1}{c}{}      & PSNR $\uparrow$         & SSIM $\uparrow$         & LPIPS $\downarrow$        & PSNR $\uparrow$       & SSIM $\uparrow$      & LPIPS $\downarrow$     & PSNR $\uparrow$          & SSIM $\uparrow$         & LPIPS $\downarrow$        & PSNR $\uparrow$       & SSIM $\uparrow$       & LPIPS $\downarrow$     \\ \midrule[0.08em]
    FLARE~\citep{zhang2025flare}                   & 16.33        & 0.574        & 0.410         & 15.35      & 0.516     & 0.591     & -             & -            & -            & -          & -          & -         \\
    AnySplat~\citep{jiang2025anysplat}                & 17.62        & 0.616        & 0.242        & 18.31       & 0.569     & 0.258     & 19.96         & 0.718        & 0.234        & 18.40      & 0.602      & 0.286      \\ \midrule
    WorldMirror                    & 20.62        & 0.706        & 0.187        & 20.92      & 0.667     & 0.203     & 25.14         & 0.859         & 0.109        & 21.25      & 0.703      & 0.223     \\
    WorldMirror (w/ intrinsics)       & \underline{22.03}        & \underline{0.765}        & \underline{0.165}        & \underline{22.08}      & \underline{0.723}     & \underline{0.175}     & \underline{25.71}         & \underline{0.877}        & \textbf{0.101}        & \underline{21.55}      & \underline{0.731}      & \underline{0.207}     \\
    WorldMirror (w/ camera pose)      & 20.84        & 0.713        & 0.182        & 21.18      & 0.674     & 0.197      & 25.14         & 0.865        & \underline{0.107}         & 21.28      & 0.700      & 0.222     \\
    WorldMirror (w/ intrinsics/camera pose)           & \textbf{22.30}        & \textbf{0.774}        & \textbf{0.155}        & \textbf{22.15}       & \textbf{0.726}     & \textbf{0.174}     & \textbf{25.77}         & \textbf{0.879}         & \textbf{0.101}        & \textbf{21.66}      & \textbf{0.736}      & \textbf{0.204}     \\ 
    \bottomrule[0.14em]
    \end{tabular}

}
\label{tab:novel_view_synthesis1}
\end{table*}